\documentclass[10pt,twocolumn,letterpaper]{article}

\usepackage[pagenumbers]{cvpr} 

\usepackage{booktabs}
\usepackage{multirow}
\usepackage{array}

\selectfont        

\def\camera{1}

\usepackage{xcolor}

\newif\ifFirstSM
\FirstSMtrue

\ifx\camera\undefined
  \newcommand{\SM}[1]{%
    \ifFirstSM
      \FirstSMfalse
      {Supplementary Material (\textit{\textcolor{cyan!50!blue}{SM}})}%
    \else
      {\textit{\textcolor{cyan!50!blue}{SM}}}%
    \fi
  }
\else
  \newcommand{\SM}[1]{Appendix~\ref{#1}}
\fi

\usepackage{textcomp}   
\usepackage{marvosym}

\definecolor{cvprblue}{rgb}{0.21,0.49,0.74}
\usepackage[pagebackref,breaklinks,colorlinks,allcolors=cvprblue]{hyperref}

\def\paperID{20480} 
\def\confName{CVPR}
\def\confYear{2026}

\title{VoQA: Visual-only Question Answering}

\author{%
  Jianing An\textsuperscript{1},
  Luyang Jiang\textsuperscript{1},
  Jie Luo\textsuperscript{1},
  Wenjun Wu\textsuperscript{1,2},
    Lei Huang\textsuperscript{1,2, \Letter} \\ 
  \\
  \textsuperscript{1}SKLCCSE, School of Artificial Intelligence, Beihang University, Beijing, China \\
  \textsuperscript{2}Hangzhou International Innovation Institute, Beihang University, Hangzhou, China \\
  \vspace{1em}
  \texttt{\{anjianing, luyang.jiang, luojie, wwj09315, huangleiAI\}@buaa.edu.cn}
}

\begin{document}
\maketitle 
\begingroup
  \renewcommand\thefootnote{\textsuperscript{\Letter}}
  
  \footnotetext{Corresponding author.}
\endgroup



\begin{abstract}

Visual understanding requires interpreting both natural scenes and the textual information that appears within them, motivating tasks such as Visual Question Answering (VQA). However, current VQA benchmarks overlook scenarios with visually embedded questions, whereas advanced agents should be able to \textit{see} the question without separate text input as humans. We introduce \textbf{Visual-only Question Answering (VoQA)}, where both the scene and the question appear within a single image, requiring models to perceive and reason purely through vision. This setting supports more realistic visual understanding and interaction in scenarios where questions or instructions are embedded directly in the visual scene. Evaluations under pure visual-only zero-shot, prompt-guided and OCR-assisted settings show that current models exhibit a clear performance drop compared to traditional VQA. To address this, we investigate question-alignment fine-tuning strategies designed to guide models toward interpreting the visual question prior to reasoning. Leveraging VoQA dataset together with these strategies yields robust vision-only reasoning while preserving cross-task generalization to traditional VQA, reflecting the complementary visual and textual reasoning capabilities fostered through VoQA training. The code\footnotemark[3] and data\footnotemark[4] are publicly available.

\footnotetext[3]{\url{https://github.com/AJN-AI/VoQA}}
\footnotetext[4]{\url{https://huggingface.co/datasets/AJN-AI/VoQA}}




\end{abstract}

\section{Introduction}
\label{sec:intro}

Visual understanding is a core capability in both human perception and artificial intelligence systems, including comprehension of natural scenes and interpretation of visual scenarios with embedded textual cues. Research in multimodal reasoning has largely been driven by the Visual Question Answering (VQA)~\cite{antol2015vqa, goyal2017making}  task, where a model receives an image together with an explicitly provided textual question and is required to produce an answer. This formulation has enabled systematic evaluation of visual reasoning and the development of a wide range of benchmarks~\cite{hudson2019gqa, marino2019ok, singh2019towards, lu2022learn} and models.

Fueled by advances in Large Vision-Language Models (LVLMs)~\cite{alayrac2022flamingo, achiam2023gpt, li2023blip, liu2023visual, Qwen2.5-VL}, recent models have achieved notable results on various VQA benchmarks.
However, current VQA benchmarks ignore cases where questions are embedded within the visual scene, yet advanced agents should be able to directly \textit{see} and interpret such questions without relying on separate textual input, similar to human perception.
Such situations commonly occur in natural environments, digital interfaces, and many embodied AI settings. In these cases, models must detect, interpret, and reason over visually embedded text without relying on separate textual inputs. Developing the ability to handle these scenarios allows embodied AI systems, including robots, autonomous vehicles, and graphical user interface (GUI) agents, to interact more efficiently and naturally with real-world environments.

\begin{figure}[t]
  \centering
  \includegraphics[width=1\linewidth]{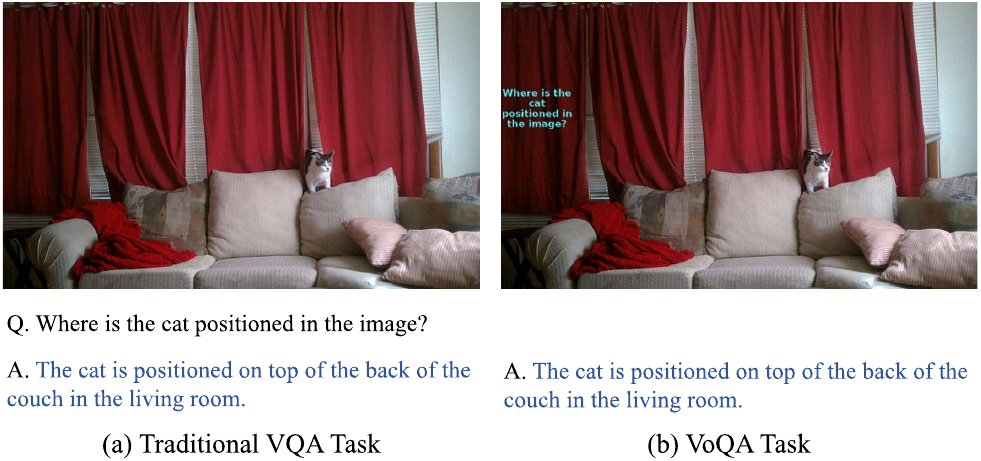}
  \caption{
  Comparison between (a) the traditional VQA task and (b) the Visual-only Question Answering (VoQA) task.
  Traditional VQA provides an image and a textual question as separate inputs, whereas VoQA embeds the question directly within the image, requiring reasoning purely through visual perception.
  }
  \label{fig:vqa_vs_voqa}

  \vspace{-10pt}
  
\end{figure}



This paper introduces \textit{Visual-only Question Answering} (VoQA), a new multimodal reasoning task that removes the explicit language channel. 
As shown in Figure~\ref{fig:vqa_vs_voqa}, the model receives a single composite image that visually encodes both the question and the scene. 
It must comprehend the embedded question, align it with the corresponding visual content, and generate the correct answer, all through a unified visual input stream. This reframing moves from explicit language-conditioned reasoning in VQA to implicit vision-only reasoning in VoQA, where the model must infer and answer questions purely from visual cues.

To support this task, we construct a large-scale \textit{VoQA Dataset} and a comprehensive \textit{VoQA Benchmark}, containing over 3.35M training and 134k evaluation samples. The training set is derived by converting LLaVA~\cite{liu2023visual} instruction-tuning data into visual-only inputs via text–image rendering. For evaluation, the benchmark is built by converting existing VQA datasets into the same visual-only format, covering tasks adapted from VQAv2, GQA, POPE~\cite{li2023evaluating}, TextVQA, and ScienceQA.

To assess model capabilities without modifying their weights, We evaluate both open-source and closed-source LVLMs under pure visual-only zero-shot, prompt-guided, and OCR-assisted experiments, revealing a substantial performance drop compared to traditional VQA. Under pure visual-only zero-shot settings, most models fail to correctly interpret and answer visually embedded questions, often merely repeating the question or even generating image caption unrelated to the embedded question. Analysis using the \textit{Question Alignment Accuracy} (QAA, see Section~\ref{sec: result analysis}) further confirms that correctly aligning visual questions is crucial, as higher QAA consistently correlates with better answer accuracy.
This finding highlights that effective reasoning depends critically on question alignment, which involves the ability to locate and interpret visual text, motivating supervised fine-tuning (SFT) strategies that guide models to first identify and align with the embedded question before producing an answer.

We conducted a comprehensive investigation of SFT strategies. For VoQA Baseline-SFT (see Section~\ref{sec:voqa-adaptation}), we found that models often either repeat the embedded question or generate answers unrelated to it, reflecting poor alignment between visual prompts and the intended response. To address this limitation, we introduce \textit{question-alignment fine-tuning strategies} that guide the model to first identify the visual question before reasoning, effectively enhancing its ability to comprehend visually embedded questions and perform reasoning purely within the visual modality.

Given the inherent difficulty of the VoQA task, it is desirable for the model to not only perform well on VoQA but also acquire \textit{cross-task generation capability} (see Section~\ref{sec: cross-task}) to traditional VQA. Among these strategies, \textit{Question–Role–Answer Supervised Fine-Tuning} (QRA-SFT, see Section~\ref{sec: qra-sft}) guides the model to predict the complete Question–Role–Answer sequence, which both enforces prior alignment with the visually embedded question, ensuring strong performance on the VoQA task, and preserves the input-output format of traditional VQA, thereby maintaining the model’s ability to generalize to text-based VQA.

\section{Related Works}

\subsection{Visual Question Answering}

Early Visual Question Answering (VQA) research focused on object recognition and basic reasoning over static images, as in VQAv1~\cite{antol2015vqa}, VQAv2~\cite{goyal2017making}, and CLEVR~\cite{johnson2017clevr}, while datasets such as Visual7W~\cite{zhu2016visual7w}, GQA~\cite{hudson2019gqa}, and VizWiz-VQA~\cite{gurari2018vizwiz} expanded to region grounding, compositional reasoning, and real-world robustness.
Subsequent work explored bias analysis~\cite{kafle2017analysis}, external knowledge~\cite{marino2019ok}, and multilingual or large-scale benchmarks such as MMBench~\cite{liu2024mmbenchmultimodalmodelallaround} and MEGA-Bench~\cite{chen2024megabench}.
To evaluate text understanding, TextVQA~\cite{singh2019towards}, OCR-VQA~\cite{mishra2019ocr}, and ChartQA~\cite{masry2022chartqa} evaluate models’ ability to read and reason over scene text or structured data.

Although these methods vary and explore different aspects of multimodal capabilities, they all provide the question as a separate textual input. In contrast, our VoQA task \textit{embeds the question as rendered text} within the image, requiring unifying perception, text recognition, and reasoning without any explicit textual prompt.

\subsection{Large Vision-Language Models}

Large Vision-Language Models (LVLMs) bridge vision and language via large-scale pretraining on image–text pairs. Early works such as CLIP~\cite{radford2021learning} and SigLIP~\cite{zhai2023sigmoid} established strong cross-modal alignment, while recent systems combine vision encoders with Large Language Models (LLMs)~\cite{achiam2023gpt, alayrac2022flamingo, li2023blip, zhu2023minigpt, liu2023visual}, enabling unified multimodal reasoning across captioning and VQA. Representative open-source models such as 
Qwen3-VL~\cite{Qwen2.5-VL}, and InternVL3.5~\cite{wang2025internvl3_5} achieve performance comparable to proprietary systems like GPT-5 and Gemini2.5-Pro.

Despite their progress, most LVLMs still depend on explicit textual input for perception and reasoning. Their reliance on language priors leaves open the question of whether multimodal understanding can emerge purely from the visual channel when textual information is embedded within the image itself.




\subsection{OCR Systems and OCR-Enhanced LVLMs}

Traditional OCR systems (e.g., Tesseract~\cite{4376991}) follow a modular detect–recognize pipeline, which limits robustness in complex or noisy scenes. Recent Transformer-based approaches such as Nougat~\cite{blecher2023nougat}, GOT-OCR2.0~\cite{wei2024general}, and DeepSeek-OCR~\cite{wei2025deepseek} pursue unified end-to-end designs, extending OCR to structured documents, formulas, and diagrams. 
Recent LVLMs demonstrate emergent OCR-like capabilities through instruction-tuned multimodal pretraining~\cite{Qwen2.5-VL, Qwen-VL, wang2025internvl3_5, zhu2025internvl3}, yet they remain dependent on vision-language instruction tuning.


In contrast, our \textit{VoQA} task removes textual input entirely, embedding questions visually within the image. This setting moves beyond text recognition, requiring models to both perceive and reason over in-image textual content purely through the visual modality.

\section{VoQA Task, Dataset and Benchmark}


We first introduce the VoQA task along with the VoQA dataset and benchmark designed for it. To investigate the model capability on this task, we evaluate both open-source and closed-source models across multiple settings.


\subsection{Task Definition}

The \textbf{Visual-only Question Answering (VoQA)} task is defined as a \emph{visual reasoning problem} in which all information required for inference, including both visual content and textual semantics, is conveyed solely through the visual modality.
Formally, let $\mathcal{I}$ denote the space of natural images, and $\mathcal{T}$ denote the space of textual sequences. In VoQA, the input to the model is a \emph{composite visual input} $I_v \in \mathcal{I}_v \subset \mathbb{R}^{H\times W\times3}$, which visually encodes both $I_s \in \mathcal{I}$ (scene image) and $T_q \in \mathcal{T}$ (question text). Given this visual-only input, a reasoning function $\mathcal{M}: \mathcal{I}_v \rightarrow \mathcal{T}_a$ produces an answer $\hat{T}_a = \mathcal{M}(I_v)$.
This formulation abstracts VoQA as a unified visual reasoning task where both the scene and the query are presented within a single visual stream, and all semantic inference is performed in the visual space without explicit textual input.

\subsection{Dataset Construction}

To build the VoQA dataset and benchmark, we introduce the text-image rendering methods used to generate visual-only inputs and the data sources used for training and evaluation.

\subsubsection{Text-Image Rendering Methods}

Given a scene image \({I}_s\) and a textual question \({T}_{q}\), to generate the composite image \({I}_{v}\) in the VoQA task, we first convert \({T}_{q} \) into RGB-format images \({I}_{q} \) of size \(224 \times 224 \) using the \textit{DejaVuSans-Bold} font. These rendered images are integrated with the original visual images via two strategies:

\textbf{(1) Concatenation}. We concatenate 
\({I}_{q} \) and \({I}_{s}\) either by resizing or padding to align their shapes, keeping textual and visual information separate. This setup serves as an intermediate task between traditional VQA and the fully integrated Watermark Rendering approach described next. Concatenation details and related experiments are provided in \SM{appendix:concatenation}. All other experiments use Watermark Rendering, which embeds questions directly within images, better reflecting real-world visual-question scenarios.

\begin{figure}[!t]
\centering
\includegraphics[width=1\linewidth]{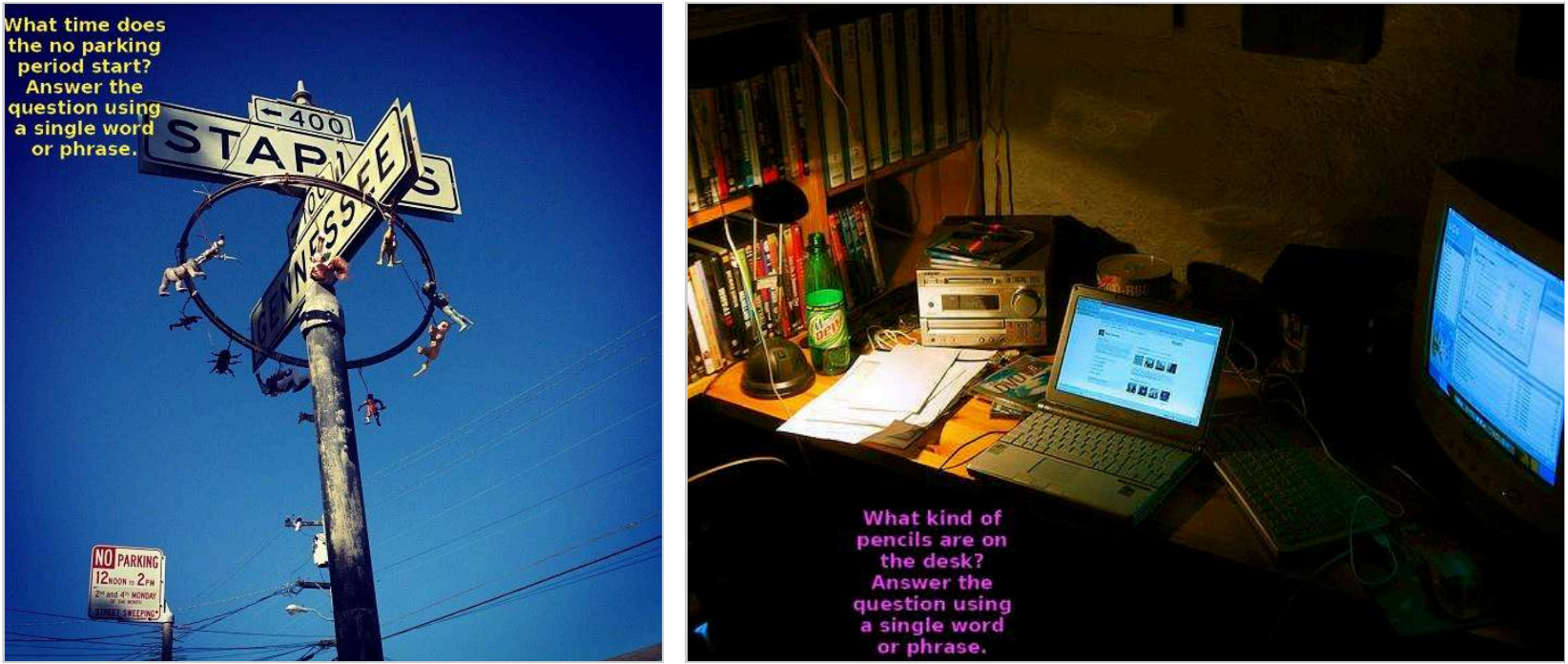}

\caption{Two examples of watermark rendering with different text colors. 
}
\vspace{-10pt}
\label{fig: examples of voqa}
\end{figure}

\textbf{(2) Watermark Rendering} (Figure~\ref{fig: examples of voqa}). To enable more integrated visual-textual understanding, we render \({I}_{q}\) as a watermark within the scene image \({I}_{s}\). Inspired by the sliding window approach, we select the embedding region based on multiple criteria 
to ensure minimal occlusion and maximal readability. Final watermark colors are selected based on WCAG contrast guidelines. Technical details, including candidate region scoring, contrast-aware color selection, are provided in \SM{appendix:watermark_rendering}.

\subsubsection{Data Preparation}

\paragraph{Training Dataset.}
We construct the VoQA training dataset based on the vision–language instruction-tuning data released by LLaVA~\cite{liu2023visual}. Each multi-turn dialogue is split into individual question–answer pairs, and each pair is rendered as a separate composite image by overlaying the question onto the corresponding scene image. The resulting dataset comprises over 3.35 million VoQA samples.

\vspace{-5pt}

\paragraph{Evaluation Benchmark.}
We build the VoQA Benchmark by transforming five widely-used VQA datasets, VQAv2~\cite{goyal2017making}, GQA~\cite{hudson2019gqa}, POPE~\cite{li2023evaluating}, TextVQA~\cite{singh2019towards}, and ScienceQA-IMG~\cite{lu2022learn} (denoted as SQA), into the visual-only format. Each image–question pair is rendered into a single composite image using our text–image rendering process. The benchmark spans a broad range of question types and reasoning skills, and includes over 134k evaluation samples. 
Details of each sub-task are provided in \SM{app:dataset imformation}.

\begin{figure*}[htbp]
\centering
\includegraphics[width=1\linewidth]{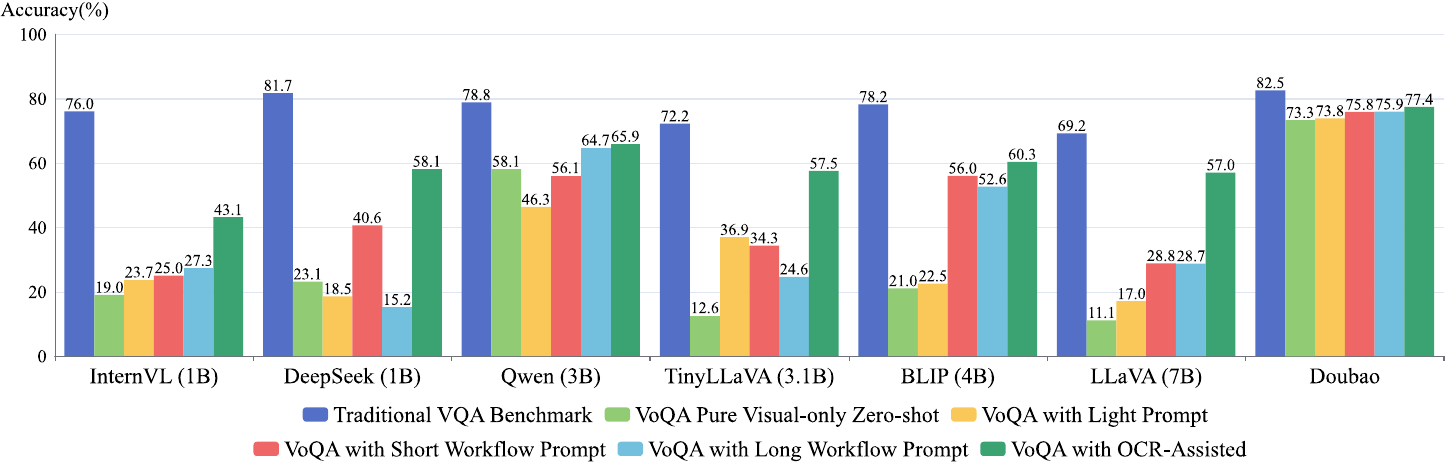}

\caption{Average Accuracy (\%) of all models on the VoQA benchmark under various zero-shot settings (\textit{pure visual-only}, \textit{prompt-guided}, and \textit{OCR-assisted}) across all datasets, compared with traditional VQA benchmarks. The models correspond to those introduced in Section~\ref{sec: evaluation setup}. Across all models and zero-shot settings, performance on VoQA is noticeably lower than on traditional VQA, highlighting the challenge of reasoning over visually embedded questions. Detailed results on each dataset are provided in \SM{appendix: zero-shot results}.}
\vspace{-10pt}
\label{fig: vqa and voqa evaluation}
\end{figure*}

\subsection{Evaluation on the VoQA Benchmark}
\label{sec: voqa evaluation}

We evaluate both open-source and closed-source models on traditional VQA benchmarks as baseline performance and on the VoQA benchmark under three main settings: (1) \textit{pure visual-only zero-shot}, (2) \textit{prompt-guided evaluation}, where models are explicitly instructed to reconstruct and answer the visually embedded question, and (3) \textit{OCR-assisted evaluation}, where a strong OCR system provides auxiliary textual input.

\subsubsection{Evaluation Setup}
\label{sec: evaluation setup}


\paragraph{Overview of Evaluation Settings.}
We evaluate seven LVLMs, including six open-source models: InternVL3-1B \cite{zhu2025internvl3}, DeepSeek-VL2-Tiny (1B) \cite{wu2024deepseek}, Qwen2.5-VL-3B-Instruct \cite{Qwen2.5-VL}, TinyLLaVA-3.1B \cite{zhou2024tinyllavaframeworksmallscalelarge}, BLIP-3 (4B) \cite{xue2024xgen}, and LLaVA-v1.5-7B \cite{liu2023visual}, as well as the closed-source model Doubao-1.5-thinking-vision-pro. 
All models are evaluated under four settings: 
(1) traditional VQA benchmarks; 
(2) the VoQA benchmark under a \textit{pure visual-only zero-shot} setting, where models receive only images with visually embedded questions; 
(3) the VoQA benchmark with \textit{prompt-guided evaluation}, where prompts explicitly instruct the model to locate and interpret the embedded question before answering; and
(4) the VoQA benchmark with \textit{OCR-assisted evaluation}, where the OCR output from a strong system (\textit{GOT-OCR 2.0}~\cite{wei2024general}) is provided as auxiliary textual input together with the composite image.


\vspace{-5pt}

\paragraph{Prompt Engineering Setup.} 
We design three prompting configurations (see \SM{app:prompts} for templates): 
(1) \textbf{Light Prompt}, which briefly instructs the model to locate and answer the embedded question; 
(2) \textbf{Short Workflow Prompt}, which adds minimal reasoning guidance and requires \textit{structured JSON outputs} containing both the detected question and final answer; 
(3) \textbf{Long Workflow Prompt}, which further enforces multi-step reasoning and stricter output constraints to ensure stronger question alignment.


To ensure fair comparison, we apply consistent response filtering across all VoQA settings. Details of the implementation are provided in \SM{appendix:responce filtering}.

\subsubsection{Evaluation Results}
\label{sec: evaluation results}

\begin{table*}[htbp]
  \centering
  \caption{A representative example from the \textit{pure visual-only} zero-shot VoQA benchmark, showing a visually embedded question and comparing responses from different models. The question \textit{"Where will these things eventually be seen? Answer the question using a single word or phrase."} is asked about a road construction site. The models correspond to those introduced in Section~\ref{sec: evaluation setup}. Note: ellipses in some model outputs are manually added for brevity and do not affect the classification of behavior type.}
  \label{tab:voqa_qual_examples}

  \resizebox{\textwidth}{!}{
  \begin{tabular}{m{5cm} l p{8.5cm} l}
    \toprule
    \textbf{Input} & \textbf{Model} & \textbf{Response} & \textbf{Behavior Type} \\
    \midrule
    \multirow{6}{*}{
  \begin{minipage}[c][5.9cm][c]{5cm}
    \centering
    \includegraphics[width=4.95cm]{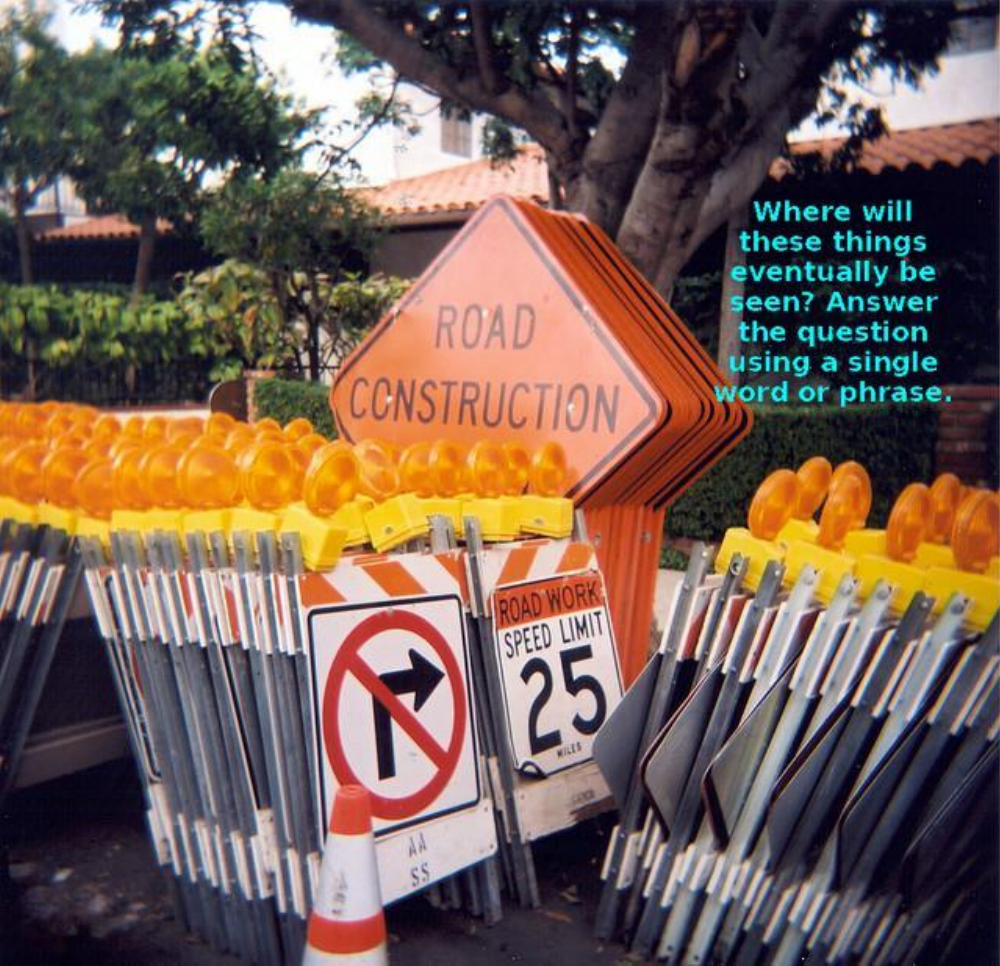}
  \end{minipage}
}
    & \multirow{2}{*}{LLaVA (7B)} & \textit{The image features a construction site with several road signs and barricades. There are ...} & \multirow{2}{*}{Question-unaware image captioning} \\
    \cline{2-4}
    & \multirow{3}{*}{InternVL (1B)} & \textit{The text in the image suggests that the question "where will these things eventually be? Answer ... phrase." is used to refer to road construction. The related sign ...} & \multirow{3}{*}{Question-aware image captioning} \\
    \cline{2-4}
    & \multirow{3}{*}{TinyLLaVA (3.1B)}
    & \textit{The image captures a scene of road construction ... The image also includes a text that reads "WHERE WILL THESE THINGS EVENLY BE ..."} & \multirow{3}{*}{Question-aware image captioning} \\
    \cline{2-4}
    & \multirow{2}{*}{DeepSeek (1B)} & \textit{Where will these things eventually be seen? Answer the question using a single word or phrase.} & \multirow{2}{*}{Repeating the question} \\
    \cline{2-4}
    & \multirow{2}{*}{BLIP (4B)} & \textit{The text 'Where will these things eventually be seen? Answer ... phrase.' is overlaid on the image.} & \multirow{2}{*}{Repeating the question} \\
    \cline{2-4}
    & Qwen (3B)
    & \textit{Disappeared} & Answering incorrectly \\
    \cline{2-4}
    & Doubao & \textit{road} & Answering correctly \\
    \bottomrule
  \end{tabular}}
\end{table*}

As shown in Figure~\ref{fig: vqa and voqa evaluation}, performance of all models drops significantly on \textit{pure visual-only zero-shot} VoQA compared with traditional VQA, highlighting the challenge of reasoning over visually embedded questions, even for the closed-source Doubao model. 

Under prompt-guided evaluation, models achieve moderate gains by explicitly parsing the embedded question. Incorporating OCR results, as an external auxiliary system that directly provides the model with the detected embedded question, achieves the highest accuracy among all VoQA settings. Nevertheless, performance remains noticeably below traditional VQA, indicating that current LVLMs still struggle when textual information is presented solely in the visual modality. 

Notably, the closed-source Doubao model performs relatively well, likely due to its large size and extensive pretraining, which help it interpret visual questions even without specialized guidance.

\subsubsection{Result Analysis}
\label{sec: result analysis}


\paragraph{Model behavior Analysis.} To better understand model behaviors under the \textit{pure visual-only} zero-shot setting, we categorize their responses to the VoQA task into five types:
(1) \textbf{Question-unaware image captioning}: ignores the embedded question and generates a generic caption;
(2) \textbf{Question-aware image captioning}: partially recognizes the question and produces a loosely related description;
(3) \textbf{Repeating the question}: restates the question without answering;
(4) \textbf{Answering incorrectly}: identifies the question but gives a wrong or irrelevant answer;
(5) \textbf{Answering correctly}: accurately understands and answers the embedded question.
We show a representative example in Table \ref{tab:voqa_qual_examples}, which refers to a road construction scene that requires understanding both the overall context and multiple text-rich elements such as warning signs and road symbols. In this case, most models fail to produce meaningful answers. Instead, they tend to either repeat the question or generate image descriptions without giving responses.

\vspace{-5pt}

\paragraph{Question Recognition Analysis.}
To analyze the performance gap, we measure models’ performance in recognizing the visually embedded question. We define \textit{Question Alignment Accuracy (QAA)} as:

\begin{equation}
\text{QAA} = 1 - \frac{\text{min(EditDistance}(\hat{q}, q))}{\text{len}(q)},
\end{equation}

where \(q\) is the ground-truth question, \(\hat{q}\) is the model-predicted question, and \textit{EditDistance} counts the character-level insertions, deletions, or substitutions to match the strings. This metric measures how accurately a model identifies the question before reasoning.

We compute the Question Alignment Accuracy (QAA) on the VoQA benchmark, reporting results by reasoning correctness for GQA, POPE, and SQA, and by confidence threshold (not less than 0.5) for TextVQA, excluding VQAv2 due to unavailable correctness labels. QAA is calculated under both workflow prompt. 

\begin{figure}[t!]
\centering
\includegraphics[width=1\linewidth]{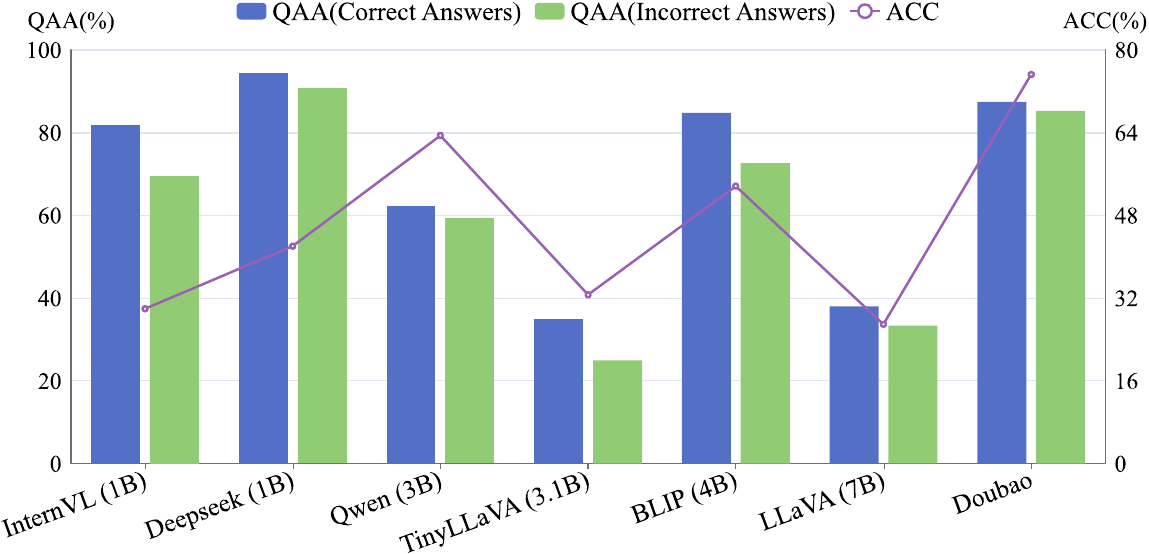}

\caption{Average Question Alignment Accuracy (QAA) and Answer Accuracy (ACC) for all models across four VoQA sub-tasks under the two workflow prompt settings, except VQAv2. \textit{Correct Answers} and \textit{Incorrect Answers} indicate averages computed over correctly and incorrectly answered samples, respectively. The models correspond to those introduced in Section~\ref{sec: evaluation setup}. For each model, the results shown correspond to the workflow setting (short or long) that yields the higher average ACC. Complete QAA and ACC results are provided in \SM{appendix: zero-shot results}.
}
\vspace{-10pt}
\label{fig: workflow}
\end{figure}

As shown in Figure~\ref{fig: workflow}, higher QAA generally correlates with stronger reasoning performance, as correctly answered samples consistently exhibit higher QAA than incorrect ones. However, a high QAA does not always guarantee higher answer accuracy (e.g., Deepseek and Qwen), likely because these models have not yet learned the reasoning pattern of first identifying the embedded question and then generating an answer based on it. Incorporating an external OCR system provides direct access to the embedded question and achieves the highest VoQA accuracy, but performance still falls short of traditional VQA due to challenges like interfering text.

Overall, these observations highlight that robust VoQA reasoning requires effective question localization and understanding, highlighting the critical role of supervised fine-tuning in teaching the model to first recognize the question and then generate the answer, as follows in the next section.

\section{Question-Alignment Fine-Tuning for VoQA}
\label{sec:qa-like sft}

\begin{figure*}[htbp]
\centering
\includegraphics[width=1\linewidth]{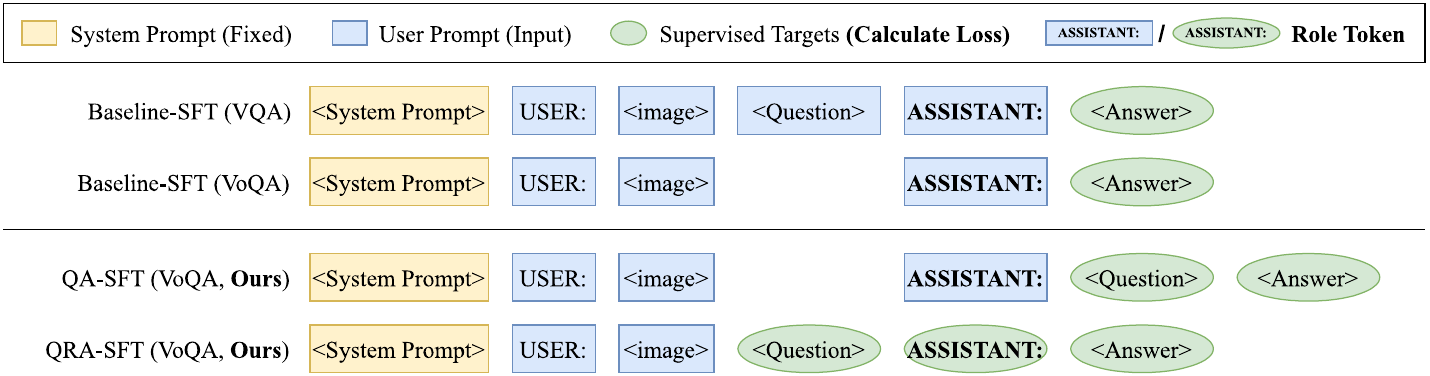}

\caption{
Comparison of four supervised fine-tuning strategies. The first two represent baseline fine-tuning under the VQA and VoQA settings, respectively. The bottom two are our proposed VoQA-specific methods that first align the visually embedded question before generating the answer.
}
\vspace{-10pt}
\label{fig: sft}
\end{figure*}

By Supervised Fine-Tuning (SFT) on instruction–response pairs from the VoQA training data, models are expected to better recognize embedded questions and reason over visual content, building on the proven effectiveness of prior SFT frameworks~\cite{dai2023instructblipgeneralpurposevisionlanguagemodels, liu2023visual, zhu2023minigpt}.
We first introduce its standard application in traditional VQA.

\vspace{-5pt}

\paragraph{VQA Baseline-SFT (denoted as VQA SFT).}
In the traditional VQA setting, the model is fine-tuned with both the image and the \textit{textual question} as inputs, using the corresponding answer as supervision (Figure~\ref{fig: sft}, Line~1). This straightforward strategy enables the model to learn visual-text reasoning conditioned on an explicitly provided question.

However, due to the difference in input formats between traditional VQA and VoQA, the VQA fine-tuning approach must be adapted when applied to the VoQA task.

\subsection{Naive SFT Adaptation to VoQA}
\label{sec:voqa-adaptation}

\paragraph{VoQA Baseline-SFT.}
In VoQA Baseline-SFT, the explicit textual question is removed, and the model generates the answer directly from a composite image \( {I}_{v}\) that visually embeds the question (Figure~\ref{fig: sft}, Line~2), preserving the basic fine-tuning structure. Subsequent experiments use three base models: \textit{TinyLLaVA-1B-Pretrained}, \textit{InternVL3-1B-Pretrained}, and \textit{Qwen2-VL-2B}. For details of experimental settings, please refer to \SM{app: sft setting}



\begin{figure}[t]
\centering
\includegraphics[width=1\linewidth]{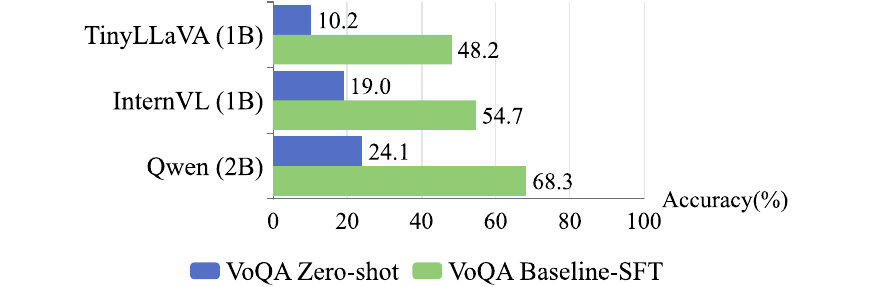}

\caption{Comparison of average accuracy on the VoQA benchmark between zero-shot evaluation and VoQA Baseline-SFT fine-tuned models. 
Detailed results for each dataset are provided in \SM{app: complete sft results}.}
\vspace{-10pt}
\label{fig: baseline}
\end{figure}

\vspace{-5pt}

\paragraph{Result Analysis.}
 As shown in Figure~\ref{fig: baseline}, VoQA Baseline-SFT moderately improves model performance over zero-shot evaluation on the VoQA task. However, we further observe that the model frequently struggles to interpret the embedded question correctly: it may repetitively restate the question without answering, or generate responses that are semantically irrelevant (See \SM{app: voqa examples} for examples).
 These behaviors reveal a fundamental limitation: without explicit textual guidance, answer-only supervision is insufficient for the model to align its instruction-following behavior with visually embedded questions.

\subsection{Enhancing Supervised Fine-Tuning with Question Alignment}
To address the limitations observed with Baseline-SFT, we introduce an explicit question alignment strategy to improve reasoning over visually embedded questions.

\vspace{-5pt}

\begin{figure}[t!]
\centering
\includegraphics[width=1\linewidth]{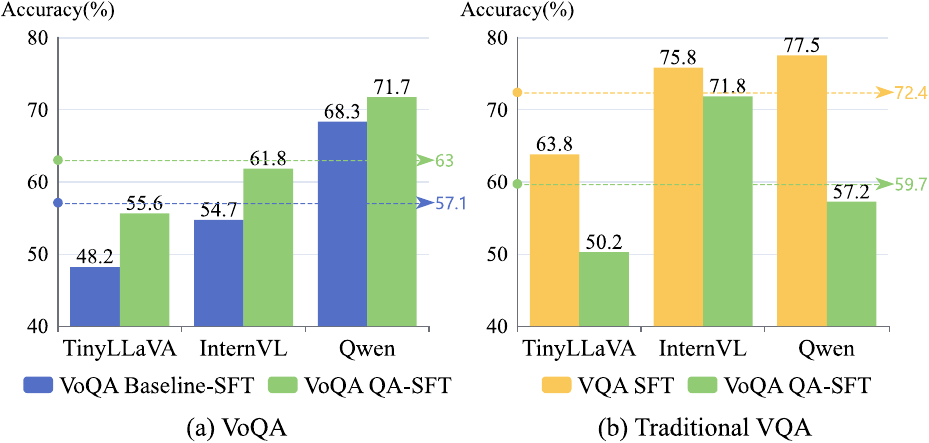}

\caption{Evaluation of models fine-tuned with different strategies. (a) Average VoQA performance of VoQA Baseline-SFT and VoQA QA-SFT models. (b) Average traditional VQA performance of VQA Baseline-SFT and VoQA QA-SFT models. All models are fine-tuned on either the LLaVA instruction-following dataset (for VQA) or the VoQA dataset (for VoQA), with differences only in input presentation. The base models correspond to the three models shown in Section~\ref{sec:voqa-adaptation}. Horizontal lines denote the mean performance across three models under each setting. Detailed results for each dataset are provided in \SM{app: complete sft results}. 
}
\vspace{-8pt}

\label{fig: qa}
\end{figure}


\paragraph{QA-SFT.}
We propose \textit{Question–Answer Supervised Fine-Tuning} (\textit{QA-SFT}, Figure~\ref{fig: sft}, Line~3), which incorporates a question reconstruction objective, supervising both the reconstructed question and its answer. This guides the model to first extract the embedded question before reasoning, thereby enforcing question–answer alignment and improving the connection between visual understanding and reasoning. As shown in Figure~\ref{fig: qa} (a), QA-SFT consistently improves average performance over VoQA Baseline-SFT across all models, indicating that aligning the model with the visually embedded question helps enhance reasoning.

\vspace{-5pt}

\label{sec: cross-task}
\paragraph{Cross-task Generation to Traditional VQA.}
Considering that VoQA is an inherently challenging task, we further investigate whether QA-SFT enables models to acquire \textit{cross-task generation capability}—namely, the capacity to generalize acquired reasoning skills beyond the specific training setting.
To assess this ability, we evaluate QA-SFT models on the traditional VQA task, where questions are provided in text rather than visually embedded. As VQA and VoQA share the same question–answering objective but differ in input modality, this evaluation serves as a natural test of cross-task generalization and provides a simpler setting to isolate the model’s visual understanding capability.

\begin{table}[t!]
    \centering
    \caption{Performance comparison of fine-tuning strategies on the VoQA benchmark. Results for VoQA QA-SFT and QRA-SFT are reported, with both models based on the same pre-trained backbone. All results are reported in accuracy (\%). 
    }

    \label{tab: qa-sft}
    \footnotesize
    \resizebox{\columnwidth}{!}{%
    \begin{tabular}{lcccccccc}
    \toprule
        \textbf{Base Model} & \textbf{Settings} & \textbf{VQAv2} & \textbf{GQA} & \textbf{POPE} & \textbf{TextVQA} & \textbf{SQA} & \textbf{Avg.}  \\ 
        \midrule
        \multirow{2}{*}{TinyLLaVA (1B)} 
         &  QA-SFT & \textbf{70.0} & \textbf{49.8} & \textbf{84.1} & \textbf{37.4} & 36.9 & \textbf{55.6}  \\ 
         &  QRA-SFT & 69.6 & 49.5 & 83.8 & 37.1 & \textbf{38.2} & \textbf{55.6} \\
         \midrule
        \multirow{2}{*}{InternVL (1B)} 
        &  QA-SFT & \textbf{73.2} & \textbf{53.0} & \textbf{86.6} & 53.7 & 42.5 & 61.8  \\ 
        &  QRA-SFT & 72.6 & 52.7 & 85.8 & \textbf{56.3} & \textbf{49.1} & \textbf{63.3}  \\ 
         \midrule
        \multirow{2}{*}{Qwen (2B)} 
        &  QA-SFT & \textbf{79.2} & 60.1 & 87.6 & \textbf{70.8} & \textbf{60.6} & \textbf{71.7}  \\ 
        &  QRA-SFT & 78.1 & \textbf{60.5} & \textbf{88.0} & \textbf{70.8} & 60.5 & 71.6  \\
         \bottomrule
    \end{tabular}
    }
\end{table}

As shown in Figure~\ref{fig: qa}, QA-SFT models perform significantly worse on traditional VQA than models fine-tuned under the standard VQA SFT setting, and in some cases even underperform their VoQA results. These findings indicate that while QA-SFT strengthens visual question alignment in VoQA, it exhibits limited cross-task generation capability toward traditional VQA.

\vspace{-6pt}

\paragraph{Structure Misalignment.}
Although QA-SFT enhances alignment with visually embedded questions in VoQA, its input–output structure differs from traditional VQA fine-tuning. In VQA SFT, the textual question is explicitly provided before the role token (\textit{ASSISTANT:}), giving the model a clear reasoning context. In contrast, QA-SFT treats the question as part of the generated output after the role token. This structural discrepancy may cause the model to misalign its reasoning or produce inaccurate answers when facing the traditional VQA format, leading to limited generalization to traditional VQA tasks.

\vspace{6pt}

\subsection{Enhancing Cross-Task Generalization in Question-Alignment Fine-Tuning}
\label{sec: qra-sft}

Building on question-alignment fine-tuning, we propose a strategy that explicitly separates question parsing from answer generation while preserving the traditional VQA format, ensuring strong alignment with visually embedded questions in VoQA and enhanced cross-task generalization to VQA compared with QA-SFT.

\vspace{-5pt}

\paragraph{QRA-SFT.}
We propose \textit{Question–Role–Answer Supervised Fine-Tuning} (QRA-SFT, Figure~\ref{fig: sft}, Line~4). In QRA-SFT, the model sequentially generates the predicted question, a role token, and the answer. This output design mirrors the traditional VQA input format, keeping compatibility while explicitly enforcing question parsing before answering. In addition, the output sequence follows a structured \textit{Question–Role–Answer} format, helping the model separate question interpretation from answer generation by predicting the role token during inference.

\subsubsection{Evaluation on VoQA and Cross-Task VQA Generalization}

\begin{figure}[t!]
\centering
\includegraphics[width=1\linewidth]{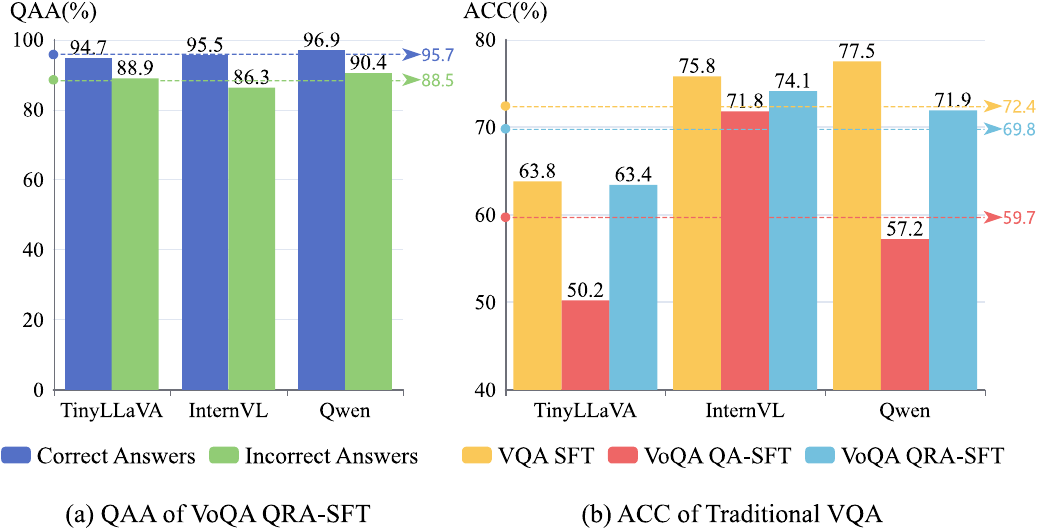}

\caption{Evaluation of models fine-tuned with different strategies.
(a) Average QAA results of VoQA QRA-SFT models on correctly and incorrectly answered samples across the four VoQA sub-tasks, excluding VQAv2.
(b) Average traditional VQA ACC of VQA Baseline-SFT, VoQA Baseline-SFT, and VoQA QA-SFT models. The fine-tuning strategies and base models are consistent with those in Figure~\ref{fig: qa}. Horizontal lines denote the mean performance across three models under each setting. Detailed results for each dataset are provided in \SM{app: complete sft results}.
}
\vspace{-8pt}
\label{fig: qra}
\end{figure}

\paragraph{Effectiveness on VoQA.}
As shown in Table~\ref{tab: qa-sft}, QRA-SFT matches or slightly outperforms QA-SFT, demonstrating that explicit supervision of question interpretation preserves alignment with visually embedded questions. Figure~\ref{fig: qra} (a) further shows that correctly answered samples consistently exhibit higher QAA than incorrect ones, exceeding 95\% on average across the three models, suggesting that strong question–answer alignment underlies the improved VoQA performance. 

\begin{figure*}[t]
    \centering
    \includegraphics[width=\linewidth]{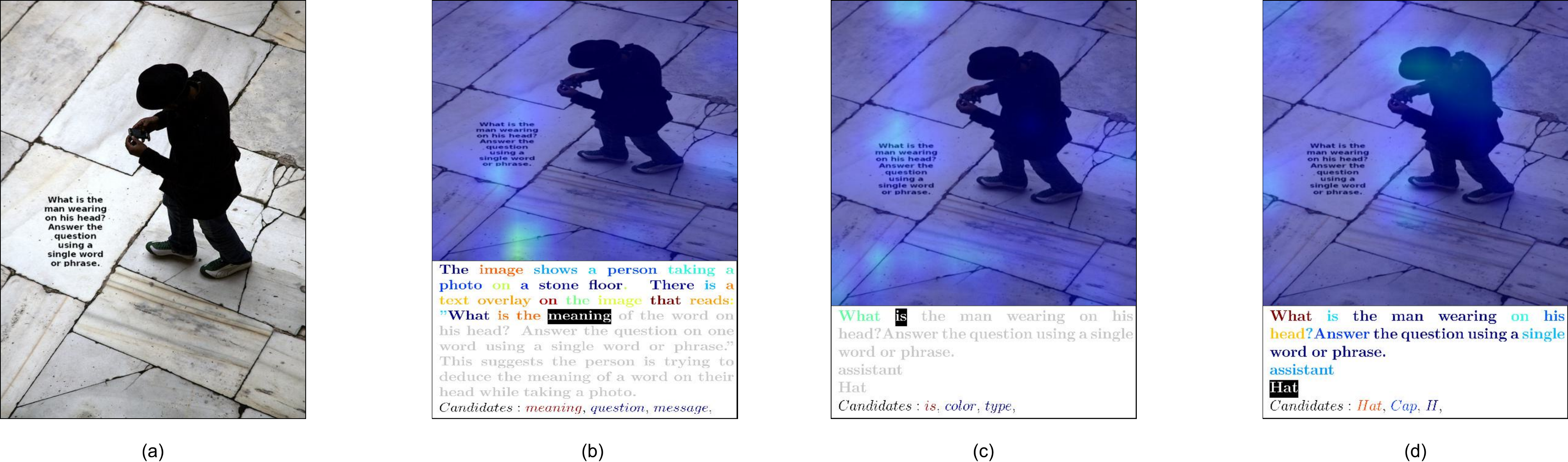}
    \caption{\textbf{Token Activation Map (TAM) visualization results.}
    (a) Original VoQA input image. 
    (b) Visualization of the \textit{InternVL3} model. 
    (c–d) Visualization of the \textit{QRA-SFT} model. 
    The QRA-SFT model first attends to the textual question region and subsequently to the relevant visual area (the hat), demonstrating a more structured reasoning process.
    \noindent\textbf{Note:} In TAM visualizations, the text below each image represents the sequence of generated tokens. The token currently being visualized is highlighted with a black box. The brightness of the overlay indicates attention strength, brighter areas correspond to higher attention on the corresponding token or image region.
    }
    \label{fig:tam_vis}

\vspace{-8pt}

\end{figure*}

\vspace{-6pt}

\paragraph{Generalization to VQA.}
Figure~\ref{fig: qra} (b) shows that models fine-tuned with QRA-SFT generalize substantially better to traditional VQA than those trained with QA-SFT. Their performance approaches that of models fine-tuned directly on VQA SFT, indicating that QRA-SFT improves question alignment in VoQA while also supporting cross-task generalization to traditional VQA.

\subsubsection{Token Activation Map Visualization}

To further illustrate how QRA-SFT improves the model’s reasoning process, we employ the \textit{Token Activation Map (TAM)} \cite{li2025tokenactivationmapvisually} technique to visualize token-level attentions on VoQA samples. Figure~\ref{fig:tam_vis} shows the comparison between the \textit{InternVL3} model and the QRA-SFT fine-tuned model. In Figure~\ref{fig:tam_vis} (a), we present the original input image, where the question text is embedded in the visual scene. Figure~\ref{fig:tam_vis} (b) shows the visualization results of the \textit{InternVL3} model. We observe that the model does not explicitly attend to the question area in the image, leading to a misunderstanding of the question and an incorrect answer. In contrast, Figures~\ref{fig:tam_vis} (c) and (d) show the \textit{QRA-SFT} model’s visualization results, where the model first focuses on the textual region containing the question, then shifts its attention to the relevant visual region (the man’s hat) when generating the final answer. These results indicate that QRA-SFT effectively guides the model to perform step-wise reasoning: first understanding the question text within the image, and then locating the visual evidence to produce the correct answer. 


\begin{table}[!t]
    \centering
    \caption{Influence of role token format and content on VoQA performance. All results are reported in accuracy (\%).}
    \label{tab: role-token}
    \resizebox{\columnwidth}{!}{%
    \begin{tabular}{lccccccc}
    \toprule
        \textbf{Model} & \textbf{Role Token} & \textbf{VQAv2} & \textbf{GQA} & \textbf{POPE} & \textbf{TextVQA} & \textbf{SQA} & \textbf{Avg.}  \\ 
        \midrule
        \multirow{2}{*}{InternVL (1B)}
        & \textit{ASSISTANT:} & \textbf{73.0} & 52.3 & \textbf{86.2} & 55.4 & \textbf{56.4} & \textbf{64.6}  \\ 
        & \textit{$\backslash$nassistant$\backslash$n} & 72.6 & \textbf{52.7} & 85.8 & \textbf{56.3} & 49.1 & 63.3  \\ 
        \midrule
        \multirow{3}{*}{TinyLLaVA (1B)} 
        & \textit{ASSISTANT:} & 69.6 & 49.5 & \textbf{83.8} & \textbf{37.1} & \textbf{38.2} & \textbf{55.6}  \\ 
        & \textit{HELPER:} & \textbf{69.7} & \textbf{49.6} & 83.7 & 36.8 & 36.9 & 55.3  \\ 
        & \textit{CAT:} & 69.4 & 49.3 & 83.6 & 36.9 & 36.6 & 55.2 \\ 
        \bottomrule
    \end{tabular}
    }
\vspace{-6pt}

\end{table}

\vspace{4pt}

\subsubsection{Ablation Study}

\paragraph{Impact of Role Token Design.}





We further analyze the effect of role token form and semantics in QRA-SFT. As shown in Table~\ref{tab: role-token}, different token formats (e.g., \textit{ASSISTANT:} vs. others) and contents (e.g., \textit{ASSISTANT} vs. \textit{HELPER}) yield nearly identical results, suggesting that the role token mainly acts as a weak structural separator to stabilize formatting rather than providing semantic guidance.

\vspace{-8pt}

\begin{figure}[!t]
\centering
\includegraphics[width=1\linewidth]{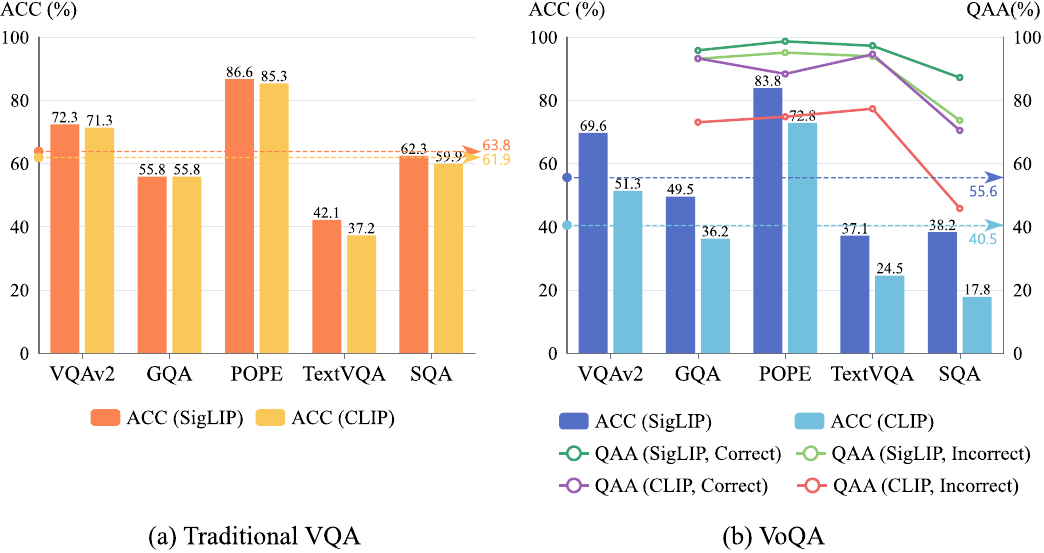}

\caption{Impact of visual encoders on VQA and VoQA tasks. \textit{TinyLLaVA-1B-Pretrained} originally uses \textbf{SigLIP} (siglip-so400m-patch14-384) as its vision encoder. We also pretrain it with \textbf{CLIP} (clip-vit-large-patch14-336) under identical fine-tuning settings. We report ACC on both VQA and VoQA, and QAA on VoQA. Horizontal lines denote the mean performance across all sub-tasks under each setting.}
\vspace{-5pt}
\label{fig: vt}
\end{figure}

\paragraph{Impact of Vision Encoders on VoQA.}
We compare two vision encoders under identical fine-tuning pipelines to isolate their effect on performance and visual–text alignment. For each vision encoder, the corresponding model is fine-tuned separately on the LLaVA instruction-tuning dataset (for VQA SFT) and the VoQA dataset. As shown in Figure~\ref{fig: vt}, both encoders perform similarly on traditional VQA, indicating comparable general visual features. On VoQA, however, SigLIP outperforms CLIP by over 15\%, emphasizing the importance of stronger visual grounding and text-sensitive representations for interpreting embedded questions. Consistently, models with higher QAA on VoQA also achieve better accuracy, confirming that visual–text alignment remains a main bottleneck in this task.

\section{Conclusion}
\label{sec:ccl}

In this work, we present \textit{VoQA}, a vision-only reasoning task designed to advance robust understanding and interaction in real-world visual scenes where textual questions are embedded directly in the image. To support this task, we build the \textit{VoQA Dataset} and \textit{VoQA Benchmark}, revealing a substantial gap between human perception and current multimodal models under vision-only conditions. Furthermore, our study of question-alignment fine-tuning demonstrates that QRA-SFT not only markedly enhances VoQA performance but also preserves strong \textit{cross-task generalization} to traditional VQA tasks. Together, these contributions establish VoQA as a practical and insightful testbed for future research on unified visual reasoning.

\noindent\textbf{Limitations and Future Work.}
Our experiments are limited to small-scale LVLMs due to computational constraints, though the proposed strategies are model-agnostic and readily scalable. 
Future directions include applying QRA-SFT to larger backbones, exploring diverse visual text styles and multilingual scenarios, and extending VoQA to embodied or interactive environments. 

\section*{Acknowledgments}
This work was partially supported by the  National Science and Technology Major Project (2022ZD0116310), National Natural Science Foundation of China (Grant No. 62476016 and  62441617), the Fundamental Research Funds for the Central Universities.

{
    \small
    \bibliographystyle{ieeenat_fullname}
    \bibliography{main}
}

\clearpage
\maketitlesupplementary

\appendix

\renewcommand{\thefigure}{A\arabic{figure}}
\renewcommand{\thetable}{A\arabic{table}}
\setcounter{figure}{0}
\setcounter{table}{0}
\setcounter{footnote}{4}

\section{Dataset Construction}


\subsection{Concatenation Methods and Results}
\label{appendix:concatenation}

\paragraph{Concatenation Process.}
Following CLIPPO \cite{tschannen2023clippo}, we first try to concatenate the scene image \({I}_{s}\) with the rendered text image \({I}_{q}\).
Considering the difference of side lengths between \({I}_{s}\) and \({I}_{q}\), we use two concatenation methods with different effects, and also take into account the four relative positions of the textual question on the top, bottom, left and right of the composite image.

\begin{figure}[!t]
\centering
\includegraphics[width=1\linewidth]{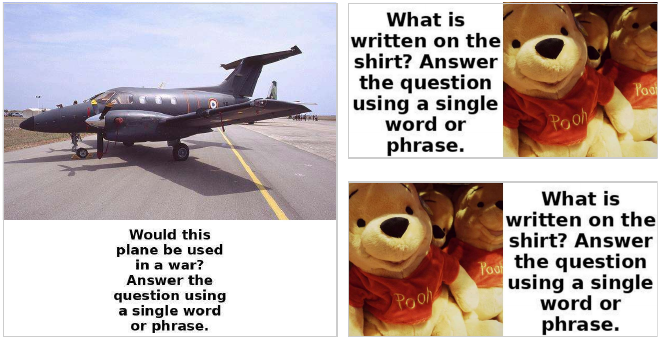}

\caption{Examples of concatenation methods, showing bottom concatenation without resizing (left) and left / right concatenation with resizing (right).}

\label{fig: examples of concatenation}
\end{figure}

 For the concatenation method without resizing (Figure~\ref{fig: examples of concatenation}, left), we align the center of the two images and fill in the blank space with the padding of the white background. For the concatenation method with resizing (Figure~\ref{fig: examples of concatenation}, right), we fix the larger size in \({I}_{s}\) and \({I}_{q}\), and enlarge the smaller size to the same side length.

\paragraph{Experiments.}

We first explored concatenation as a simple baseline, which performed better than watermark rendering thanks to its fixed layout and clean background. 
We averaged results across four concatenation directions for each method and evaluated model performance in three settings: the traditional VQA task, the \textit{pure visual-only} VoQA task, and the VoQA task with Light Prompt (as described in Section~\ref{app:prompts}). Results are shown in Table~\ref{tab: concat without resize} and Table~\ref{tab: concat with resize}. Even under this simplified setup, where the scene image and the embedded question are completely separated, models still struggle to perform well. The effect of Light Prompt also varies across models, sometimes improving but sometimes reducing performance. Overall, both concatenation methods outperform watermark rendering in the pure visual-only setting, consistent with their simpler visual composition.

\subsection{Details of Watermark Rendering Method}
\label{appendix:watermark_rendering}

We first identify the most suitable region for rendering the watermark to minimize the impact of poor text readability. The watermark's side length is set to 1/4 of the short side of the scene image \({I}_{s}\). To balance efficiency and quality, candidate regions are generated using a stride of 1/4 the image side length. Each region is scored based on gradient, variance, and contrast (weighted 0.4, 0.4, and 0.2, respectively), reflecting smoothness and color contrast to ensure readability.

Next, we determine the watermark color in HSV space based on the average hue, saturation, and value of the selected region. The hue is set to the complementary hue (offset by 180° on the color wheel). If the saturation exceeds 200, the watermark saturation is set to 0.8 times that of the region; otherwise, it is set to the maximum value of 255. The value (brightness) is inverted: if the region’s value exceeds 127, the watermark value is set to 0; otherwise, 255.

Finally, we render \({I}_{q}\) onto \({I}_{s}\). Following the WCAG standard\footnote{\url{https://www.w3.org/Translations/WCAG21-zh/}}, if the contrast ratio between the watermark and background exceeds 4.5, we retain the computed color; otherwise, we choose either black or white, selecting the one with greater contrast.

\subsection{Details of VoQA Benchmark}
\label{app:dataset imformation}


We remove reference OCR tokens from the TextVQA questions to prevent excessively long text from reducing font size and interfering with the question information already present in the image. To ensure visual clarity, we filter out overly long questions in SQA to avoid unreadable text regions. 

Our benchmark includes three question types: open-ended questions (VQAv2, GQA, TextVQA), binary questions (POPE), and multiple-choice questions (SQA, with A/B/C/D options embedded directly in the images). The final number of evaluation samples retained for each dataset is 107,394 (VQAv2), 12,578 (GQA), 8,910 (POPE), 5,000 (TextVQA), and 1,087 (SQA).

\clearpage
\begin{table*}[ht!]
    \centering
    \caption{Evaluation results of \textit{concatenation method without resizing}. We evaluate four  open-source models under three settings—the traditional VQA task, the \textit{pure visual-only} VoQA task, and the VoQA task with Light Prompt. Each result is averaged over four concatenation directions. All results are reported in accuracy (\%). Although the image content and embedding questions are presented separately, which helps maintain visual consistency, the model's performance still drops significantly compared to the traditional VQA setup, highlighting the inherent difficulties of VoQA.}
    \label{tab: concat without resize}
    \footnotesize
    \begin{tabular}{ll cccccc}
    \toprule
        \textbf{Model} & \textbf{Settings} & \textbf{VQAv2} & \textbf{GQA} & \textbf{POPE} & \textbf{TextVQA} & \textbf{SQA} & \textbf{Avg.}   \\ 
        \midrule
        \multirow{3}{*}{InternVL2.5-1B} 
        & Traditional VQA & \textbf{76.3} & \textbf{56.9} & \textbf{89.9} & \textbf{72.0} & \textbf{96.7} & \textbf{78.4} \\
        & VoQA zero-shot & 4.1 & 3.0 & 54.9 & 15.3 & 4.1 & 16.3   \\ 
        & Light Prompt & \underline{35.8} & \underline{20.1} & \underline{65.8} & \underline{43.7} & \underline{17.5} & \underline{36.6}   \\ 
        \midrule
        \multirow{3}{*}{DeepSeek-VL2-Tiny} 
        & Traditional VQA & \textbf{83.5} & \textbf{62.7} & \textbf{88.6} & \textbf{79.2} & \textbf{94.3} & \textbf{81.7} \\
        & VoQA zero-shot & \underline{9.3} & \underline{4.5} & \underline{71.1} & \underline{31.2} & \underline{40.3} & \underline{31.3}   \\ 
        & Light Prompt & 8.3 & 4.0 & 54.0 & 4.3 & 6.7 & 15.5   \\ 
        \midrule
        \multirow{3}{*}{Qwen2.5-VL-3B-Instruct}
        & Traditional VQA & \textbf{82.0} & \textbf{60.1} & \textbf{88.2} & \textbf{79.0} & \textbf{84.9} & \textbf{78.8} \\
        & VoQA zero-shot & \underline{70.7} & \underline{47.9} & \underline{81.2} & \underline{65.4} & \underline{69.2} & \underline{66.9}   \\ 
        & Light Prompt & 25.1 & 15.9 & 80.9 & 38.6 & 32.8 & 38.6   \\ 
        \midrule
        \multirow{3}{*}{BLIP-3 (4B)}
        & Traditional VQA & \textbf{81.7} & \textbf{61.6} & \textbf{87.0} & \textbf{71.0} & \textbf{89.8} & \textbf{78.2} \\
        & VoQA zero-shot & 10.1 & 5.8 & 53.6 & \underline{16.4} & \underline{26.2} & \underline{22.4}   \\ 
        & Light Prompt & \underline{11.7} & \underline{8.6} & \underline{61.5} & 15.6 & 6.3 & 20.8  \\ 
        \bottomrule
    \end{tabular}
\end{table*}

\begin{table*}[ht!]
    \centering
    \caption{Evaluation results of \textit{concatenation method with resizing}. Experiment settings are identical to those in Table~\ref{tab: concat without resize}, except for the difference in the concatenation method used during data construction. All results are reported in accuracy (\%). Despite the simpler setup compared to watermark rendering, all models still show a significant performance drop relative to traditional VQA.}
    \label{tab: concat with resize}
    \footnotesize
    \begin{tabular}{ll cccccc}
    \toprule
        \textbf{Model} & \textbf{Settings} & \textbf{VQAv2} & \textbf{GQA} & \textbf{POPE} & \textbf{TextVQA} & \textbf{SQA} & \textbf{Avg.}   \\ 
        \midrule
        \multirow{3}{*}{InternVL2.5-1B} 
        & Traditional VQA & \textbf{76.3} & \textbf{56.9} & \textbf{89.9} & \textbf{72.0} & \textbf{96.7} & \textbf{78.4} \\
        & VoQA zero-shot & 4.8 & 2.7 & 54.6 & 14.4 & 5.8 & 16.5 \\ 
        & Light Prompt & \underline{35.2} & \underline{21.0} & \underline{68.2} & \underline{42.9} & \underline{16.0} & \underline{36.6} \\ 
        \midrule
        \multirow{3}{*}{DeepSeek-VL2-Tiny} 
        & Traditional VQA & \textbf{83.5} & \textbf{62.7} & \textbf{88.6} & \textbf{79.2} & \textbf{94.3} & \textbf{81.7} \\
        & VoQA zero-shot & 11.0 & 5.6 & \underline{69.9} & \underline{36.4} & \underline{40.7} & \underline{32.7}  \\ 
        & Light Prompt & \underline{11.9} & \underline{6.6}& 57.4 & 3.9 & 6.8 & 17.3  \\ 
        \midrule
        \multirow{3}{*}{Qwen2.5-VL-3B-Instruct}
        & Traditional VQA & \textbf{82.0} & \textbf{60.1} & \textbf{88.2} & \textbf{79.0} & \textbf{84.9} & \textbf{78.8} \\
        & VoQA zero-shot & \underline{69.2} & \underline{45.4} & \underline{80.6} & \underline{60.9} & \underline{72.7} & \underline{65.8}  \\ 
        & Light Prompt & 24.1 & 15.2 & 80.2 & 30.4 & 47.6 & 39.5  \\ 
        \midrule
        \multirow{3}{*}{BLIP-3 (4B)}
        & Traditional VQA & \textbf{81.7} & \textbf{61.6} & \textbf{87.0} & \textbf{71.0} & \textbf{89.8} & \textbf{78.2} \\
        & VoQA zero-shot & 10.6 & 7.6 & 53.6 & \underline{17.7} & \underline{26.6} & \underline{23.2} \\ 
        & Light Prompt & \underline{15.3} & \underline{10.3} & \underline{55.6} & 16.4 & 7.9 & 21.1 \\ 
        \bottomrule
    \end{tabular}
\end{table*}

\clearpage  

\section{LVLM Evaluation Settings and Results}
\label{appendix:data evaluation}


\subsection{Examples of Designed Prompts}
\label{app:prompts}

\paragraph{Light Prompt.}
We explore a series of carefully designed light prompts to guide the model in performing the VoQA task under a zero-shot setting. By analyzing the impact of different key phrases in the prompts, we identify the most effective formulation. Table~\ref{tab:several_prompts} summarizes the various prompt designs and highlights their key instructional components.

\begin{table*}[htbp]
\centering
\caption{Light prompt design variants for zero-shot VoQA. Each prompt aims to instruct the model to first locate the embedded question in the image and then generate an answer based solely on the visual content. Key phrases used in the prompts are listed to highlight their instructional focus. The first prompt is the one ultimately adopted for our light prompt evaluations.}
\label{tab:several_prompts}
\footnotesize
\begin{tabular}{m{9cm} m{4cm}}
\toprule
\textbf{Designed prompts} & \textbf{Key phrases} \\
\midrule
\textbf{There is a question in this image, you need to find the question and answer the question based on the visual information of the entire image. (Ours)} & find the question, answer the question, the visual information \\
\midrule
There is a question in this image. You need to find the question and answer the question based on the visual information of the entire image. Please do not repeat the question and answer it directly. & find the question, answer the question, the visual information, do not repeat the question, answer it directly \\
\midrule
Please answer the question in the image directly, and do not repeat the question. You need to find the answer based on the visual information of the entire image. & answer the question directly, do not repeat the question, the visual information\\
\midrule
Please answer the question in the image directly, and do not repeat the question. & answer the question, do not repeat the question \\
\midrule
Please find the question in the image, and answer it directly based on the whole image. & find the question, answer it directly \\
\midrule
Please find the question in the image, and answer it directly based on the whole image. Do not repeat the question. & find the question, answer it directly, do not repeat the question \\
\midrule
Please find the question in the image and answer it based on the image. Do not repeat the question you find in your answers. & find the question, answer it, do not repeat the question \\

\bottomrule
\end{tabular}
\end{table*}

\paragraph{Short Workflow Prompt.}
This prompt guides the model through image analysis, ensuring it extracts the question, understands its context, and provides a precise answer based on visual cues. The placeholders \texttt{<bbox>}, \texttt{<top-left-location>}, \texttt{<bottom-right-location>}, \texttt{<picture-width>} and \texttt{<picture-height>} are replaced by the actual question location and image dimensions, derived from the dataset generation process. The model is required to output the result in a strict JSON format, ensuring consistency and enabling automated processing. Prompt content is as follows:

\texttt{\\
Task Definition:\\
You will receive an image with a watermark question. Your task is to:\\
1. Detect and extract the full question text.\\
2. Locate the question bounding box <bbox> (top-left <top-left-location>, bottom-right <bottom-right-location>) in the <picture-width>x<picture-height> image.\\
3. Understand the question type.\\
4. Answer accurately based only on the visual content.\\
\\
Output Format (strict JSON):\\
\{\\
    "Detected Question": "<recognized question text>",\\
    "Answer": "<concise answer based on the image>",\\
    "Reasoning": "<brief explanation of how the answer was derived>"\\
\}\\
\\
Important:\\
- Ensure full and coherent question extraction.\\
- Base the answer strictly on visual evidence.\\
- If uncertain or unclear, output "Unknown".\\
- Do not add commentary outside the JSON.\\
\\
Now, analyze the image, detect the question and its location, and output the result in the required JSON format.
}

\paragraph{Long Workflow Prompt.}
The Long Workflow Prompt offers a more detailed step-by-step process compared to the Short Workflow Prompt. It includes tasks like watermark question detection, visual information extraction, and answer generation. However, both workflow prompts share the same setup regarding placeholders, such as \texttt{<bbox>}, \texttt{<top-left-location>}, \texttt{<bottom-right-location>}, \texttt{<picture-width>} and \texttt{<picture-height>}, which are replaced by the actual question location and image dimensions based on the dataset generation process.

While the Short Workflow Prompt is more concise, the Long Workflow Prompt provides more structured guidance, ensuring that the model follows a clear workflow for accurate answers based solely on the visual content. Prompt content is as follows:

\texttt{\\
(1) Task Definition:\\
You will receive an image containing a watermark-embedded question. Your task is to:\\
1. Detect the full question text embedded as a watermark in the image.\\
2. Understand the meaning of the question.\\
3. Answer the question based solely on the visual content of the image.\\
4. Provide an accurate and relevant answer.\\
\\
(2) Workflow Steps:\\
Step 1: Watermark Question Detection\\
- Scan the image for textual content, including semi-transparent overlays and repeated patterns.\\
- Extract the complete question sentence. \\
- Tip: The question is located at bounding box <bbox>, from top-left <top-left-location> to bottom-right <bottom-right-location> in a <picture-width>x<picture-height> image.\\
\\
Step 2: Visual Information Extraction\\
- Analyze the image to find information relevant to the question.\\
- Focus on object recognition, counting, attribute description, spatial relations, scene understanding, text recognition, or reasoning as needed.\\
\\
Step 3: Answer Generation\\
- Provide an accurate, concise answer based solely on visual evidence.\\
- Give a brief explanation of how the answer was derived.\\
- If the answer cannot be determined, state it honestly.\\
\\
(3) Output Format (JSON):\\
Strictly return a valid JSON object as shown below:\\
\{\\
    "Detected Question": "<recognized question text>",\\
    "Answer": "<concise answer based on the image>",\\
    "Reasoning": "<brief explanation of how the answer was derived>"\\
\}\\
\\
Example:\\
\{\\
    "Detected Question": "What is the brand of this camera?",\\
    "Answer": "Canon", \\
    "Reasoning": "The text 'Canon' is clearly visible on the camera body."\\
\}\\
\\
(4) Important Notes:\\
1. Ensure question detection and answer are grounded in the image.\\
2. Provide the full, coherent question text.\\
3. Base the answer strictly on visual evidence.\\
4. Be concise and clear.\\
5. State uncertainty clearly if the image lacks information.\\
\\
Now, process the input image and execute the VoQA task following the above workflow.
}

\begin{table*}[!t]
\centering
\caption{
Performance comparison between traditional VQA and VoQA under several zero-shot settings for VoQA. The evaluation includes six open-source models and two closed-source models. All results are reported in accuracy (\%).
}
\label{tab: vqa vs voqa}
\footnotesize
\begin{tabular}{lllcccccc}
\toprule
\textbf{Model} & \textbf{Benchmark} & \textbf{Setting} & \textbf{VQAv2} & \textbf{GQA} & \textbf{POPE} & \textbf{TextVQA} & \textbf{SQA} & \textbf{Avg.} \\ \midrule

\multirow{6}{*}{InternVL3-1B} 
 & Traditional VQA & / & \textbf{71.8} & \textbf{52.1} & \textbf{88.8} & \textbf{71.8} & \textbf{95.4} & \textbf{76.0} \\
 \cline{2-9}
 & \multirow{5}{*}{VoQA} & pure zero-shot & 5.5 & 3.2 & 53.9 & 12.8 & 19.5 & 19.0 \\
 & & light prompt & 14.6 & 9.2 & 57.8 & 19.0 & 18.0 & 23.7  \\ 
 & & short workflow prompt & 13.6 & 7.6 & 53.6 & 26.5 & 23.6 & 25.0  \\ 
 & & long workflow prompt & 16.8 & 10.0 & 59.9 & 30.8 & 18.9 & 27.3  \\
 & & OCR-assisted & \textbf{45.3} & \textbf{29.4} & \textbf{63.3} & \textbf{31.5} & \textbf{45.7} & \textbf{43.0} \\ 
 
\midrule

\multirow{6}{*}{DeepSeek-VL2-Tiny (1B)} 
 & Traditional VQA & / & \textbf{83.5} & \textbf{62.7} & \textbf{88.6} & \textbf{79.2} & \textbf{94.3} & \textbf{81.7} \\
 \cline{2-9}
 & \multirow{5}{*}{VoQA} & pure zero-shot & 2.0 & 0.2 & 59.3 & 21.7 & 32.0 & 23.1 \\
 & & light prompt & 7.5 & 3.7 & 57.0 & 6.4 & 18.0 & 18.5  \\ 
 & & short workflow prompt & 35.1 & 21.5 & 58.5 & \textbf{48.7} & 39.3 & 40.6  \\ 
 & & long workflow prompt & 8.3 & 5.2 & 53.1 & 4.9 & 4.5 & 15.2  \\ 
 & & OCR-assisted & \textbf{69.1} & \textbf{44.7} & \textbf{82.3} & 41.3 & \textbf{52.9} & \textbf{58.1}  \\
\midrule

\multirow{6}{*}{Qwen2.5-VL-3B-Instruct} 
 & Traditional VQA & / & \textbf{82.0} & \textbf{60.1} & \textbf{88.2} & \textbf{79.0} & \textbf{84.9} & \textbf{78.8} \\
  \cline{2-9}
 & \multirow{5}{*}{VoQA} & pure zero-shot & 57.4 & 35.0 & 76.3 & 60.7 & 61.2 & 58.1 \\
 & & light prompt & 40.1 & 26.9 & 71.0 & 37.1 & 56.2 & 46.3  \\ 
 & & short workflow prompt & 57.0 & 36.9 & 55.7 & 64.9 & 66.2 & 56.1  \\ 
 & & long workflow prompt & \textbf{69.9} & 45.7 & 80.0 & \textbf{66.6} & 61.2 & 64.7  \\ 
 & & OCR-assisted & 68.0 & \textbf{47.8} & \textbf{84.2} & 58.7 & \textbf{70.9} & \textbf{65.9} \\ 
\midrule

\multirow{6}{*}{TinyLLaVA-3.1B} 
 & Traditional VQA & / & \textbf{80.1} & \textbf{62.1} & \textbf{87.2} & \textbf{55.9} & \textbf{75.5} & \textbf{72.2} \\
   \cline{2-9}
 & \multirow{5}{*}{VoQA} & pure zero-shot & 0.8 & 0.2 & 50.5 & 4.5 & 6.7 & 12.6 \\
 & & light prompt & 46.0 & 22.2 & \textbf{71.6} & 31.8 & 12.7 & 36.9  \\ 
& & short workflow prompt & 41.4 & 25.0 & 65.2 & 28.3 & 11.9 & 34.3  \\ 
& & long workflow prompt & 28.8 & 16.2 & 59.9 & 12.3 & 6.0 & 24.6  \\
& & OCR-assisted & \textbf{70.0} & \textbf{51.8} & 71.4 & \textbf{46.3} & \textbf{47.9} & \textbf{57.5} \\    
\midrule

\multirow{6}{*}{BLIP-3 (4B)} 
 & Traditional VQA & / & \textbf{81.7} & \textbf{61.6} & \textbf{87.0} & \textbf{71.0} & \textbf{89.8} & \textbf{78.2} \\
 \cline{2-9}
 & \multirow{5}{*}{VoQA} & pure zero-shot & 7.3 & 2.7 & 49.7 & 16.2 & 29.1 & 21.0 \\
 & & light prompt & 11.5 & 13.3 & 58.0 & 11.3 & 18.8 & 22.5  \\ 
 & & short workflow prompt & 65.7 & 42.7 & \textbf{80.1} & 54.3 & 37.4 & 56.0  \\ 
 & & long workflow prompt & 63.9 & 40.3 & 79.6 & 50.7 & 28.6 & 52.6 \\ 
 & & OCR-assisted & \textbf{68.6} & \textbf{46.4} & 79.3 & \textbf{57.1} & \textbf{50.1} & \textbf{60.3} \\
\midrule

\multirow{6}{*}{LLaVA-v1.5-7B} 
 & Traditional VQA & / & \textbf{78.5} & \textbf{62.0} & \textbf{85.9} & \textbf{46.1} & \textbf{73.7} & \textbf{69.2} \\
\cline{2-9}
 & \multirow{5}{*}{VoQA} & pure zero-shot & 0.2 & 0 & 50.5 & 3.5 & 1.5 & 11.1 \\
 & & light prompt & 13.7 & 9.3 & 55.0 & 2.9 & 4.1 & 17.0  \\ 
 & & short workflow prompt & 36.4 & 18.4 & 60.3 & 22.6 & 6.2 & 28.8  \\ 
 & & long workflow prompt & 38.1 & 18.4 & 62.1 & 21.5 & 3.6 & 28.7  \\ 
 & & OCR-assisted & \textbf{71.9} & \textbf{55.1} & \textbf{74.7} & \textbf{44.2} & \textbf{39.0} & \textbf{57.0} \\
\midrule

\multirow{6}{*}{Doubao-1.5-thinking-vision-pro} 
 & Traditional VQA & / & \textbf{82.7} & \textbf{61.7} & \textbf{89.1} & \textbf{82.1} & \textbf{96.9} & \textbf{82.5} \\ 
\cline{2-9}
 & \multirow{5}{*}{VoQA} & pure zero-shot & 75.7 & 49.1 & 86.2 & 78.1 & 77.6 & 73.3 \\
 & & light prompt & 77.0 & 49.7 & 86.2 & 78.2 & 77.6 & 73.8  \\ 
 & & short workflow prompt & 78.3 & 52.5 & 86.6 & 79.0 & 82.3 & 75.8 \\
 & & long workflow prompt & 78.9 & 53.3 & 86.5 & \textbf{79.5} & 81.2 & 75.9 \\ 
 & & OCR-assisted & \textbf{79.5} & \textbf{55.1} & \textbf{87.7} & 78.8 & \textbf{86.0} & \textbf{77.4} \\
\midrule

\multirow{2}{*}{GPT-4o} 
 & Traditional VQA & / & \textbf{72.5} & \textbf{51.3} & \textbf{86.5} & \textbf{75.2} & \textbf{90.7} & \textbf{75.2} \\
 \cline{2-9}
 & VoQA & pure zero-shot & 70.5 & 48.3 & 85.7 & 69.4 & 78.3 & 70.4 \\
 
\bottomrule

\end{tabular}
\label{tab:voqa_vs_vqa}
\end{table*}

\subsection{Response Filtering Method}
\label{appendix:responce filtering}

To fairly evaluate model performance on the VoQA Benchmark, we apply standardized response filtering, since models may produce answers in varying formats, while the benchmark datasets require concise responses, typically a single letter, word, or phrase.

For zero-shot and few-shot (see section~\ref{app:few-shot-settings}) models, we adopt multiple parsing strategies to extract the actual answer from the response. These include formats such as \texttt{The answer is [Actual Answer]} or \texttt{Answer: [Actual Answer]}. Dataset-specific rules are also considered. For instance, POPE only evaluates whether the response contains \texttt{no} or \texttt{not}, so when the exact answer location is uncertain, we retain the full sentence to preserve contextual clues.

For experiments using JSON-formatted outputs (including long/short workflow prompt and few-shot settings), we first check whether the model's response is in valid JSON format. If so, we parse the JSON and extract the value of the \texttt{ANSWER} field as the candidate answer; otherwise, we retain the original output. Finally, we apply the multi-strategy filtering approach described above to obtain the final answer.

For fine-tuned models, filtering strategies vary depending on the fine-tuning approach. For all models fine-tuned using Baseline-SFT, we retain the original output. For QRA-SFT, we extract the segment following the last occurrence of a role token as the answer. Specifically for QA-SFT, where responses often follow a \textit{Question + Answer} format, we directly use the last sentence as the final predicted answer.

\subsection{Zero-shot Evaluation Results}
\label{appendix: zero-shot results}

\paragraph{Detailed ACC Results for Zero-shot Evaluation.} The complete results of the traditional VQA benchmarks and the VoQA zero-shot evaluation across different settings are summarized in Table~\ref{tab: vqa vs voqa}. All sub-datasets show a clear and consistent performance degradation on VoQA, highlighting the increased complexity and distinct challenges posed by the VoQA task.

\paragraph{QAA and ACC under Workflow-based Prompt-Guided Evaluation.} As shown in Table~\ref{tab: short workflow analysis} and Table~\ref{tab: long workflow analysis}, across most sub-tasks, models exhibit higher QAA on correctly answered samples than on incorrect ones, indicating that recognizing the embedded question is a prerequisite for reliable reasoning. Moreover, in the majority of sub-tasks, higher QAA correlates with higher ACC, further reinforcing the importance of accurate question understanding.

\begin{table}[!t]
    \centering
    \caption{QAA and ACC Results under short workflow settings.}
    \label{tab: short workflow analysis}
    \resizebox{\columnwidth}{!}{%
    \begin{tabular}{ll ccccc}
    \toprule
        \textbf{Model} & \textbf{Metric} & \textbf{GQA} & \textbf{POPE} & \textbf{TextVQA} & \textbf{SQA} & \textbf{Avg.}  \\ 
        \midrule
        \multirow{3}{*}{InternVL (1B)} & QAA(Correct) & \textbf{92.4} & \textbf{87.4} & \textbf{92.6} & \textbf{66.6} & \textbf{84.7}  \\ 
        & QAA(Incorrect) & 83.8 & 87.2 & 76.1 & 55.4 & 75.6  \\ 
        \cline{2-7}
        & ACC & 7.6 & 53.6 & 26.5 & 23.6 & 27.8  \\ 
        \midrule
        \multirow{3}{*}{DeepSeek (1B)} & QAA(Correct) & \textbf{99.2} & \textbf{99.3} & \textbf{98.6} & \textbf{80.5} & \textbf{94.4}  \\ 
        & QAA(Incorrect) & 98.9 & \textbf{99.3} & 95.2 & 69.4 & 90.7  \\ 
        \cline{2-7}
        & ACC & 21.5 & 58.5 & 48.7 & 39.3 & 42.0  \\ 
        \midrule
        \multirow{3}{*}{Qwen (3B)} & QAA(Correct) & \textbf{59.4} & \textbf{61.1} & \textbf{79.0} & \textbf{53.8} & \textbf{63.3}  \\ 
        & QAA(Incorrect) & 59.1 & 57.5 & 78.3 & 47.1 & 60.5  \\ 
        \cline{2-7}
        & ACC & 36.9 & 55.7 & 64.9 & 66.2 & 55.9  \\ 
        \midrule
        \multirow{3}{*}{TinyLLaVA (3.1B)} & QAA(Correct) & \textbf{34.9} & \textbf{52.8} & \textbf{21.2} & \textbf{31.2} & \textbf{35.0}  \\ 
        & QAA(Incorrect) & 24.8 & 45.5 & 13.2 & 16.2 & 24.9  \\ 
        \cline{2-7}
        & ACC & 25.0 & 65.2 & 28.3 & 11.9 & 32.6  \\ 
        \midrule
        \multirow{3}{*}{BLIP (4B)} & QAA(Correct) & \textbf{86.4} & \textbf{93.2} & \textbf{86.3} & \textbf{73.1} & \textbf{84.7}  \\ 
        & QAA(Incorrect) & 79.1 & 91.6 & 64.4 & 55.3 & 72.6  \\ 
        \cline{2-7}
        & ACC & 42.7 & 80.1 & 54.3 & 37.4 & 53.6  \\ 
        \midrule
        \multirow{3}{*}{LLaVA (7B)} & QAA(Correct) & \textbf{38.7} & 36.0 & \textbf{37.5 }& \textbf{40.0} & \textbf{38.0} \\ 
        & QAA(Incorrect) & 35.9 & \textbf{36.7} & 34.2 & 26.3 & 33.3  \\ 
        \cline{2-7}
        & ACC & 18.4 & 60.3 & 22.6 & 6.2 & 26.9  \\ 
        \midrule
        \multirow{3}{*}{Doubao} & QAA(Correct) & \textbf{94.3} & \textbf{98.1} & \textbf{99.5} & \textbf{57.6} & \textbf{87.4}  \\ 
        & QAA(Incorrect) & 93.6 & 97.3 & 98.9 & 50.9 & 85.2  \\ 
        \cline{2-7}
        & ACC & 52.5 & 86.6 & 79.0 & 82.8 & 75.2 \\
        \bottomrule
    \end{tabular}
    }
\end{table}

\begin{table}[!t]
    \centering
    \caption{QAA and ACC Results under long workflow settings.}
    \label{tab: long workflow analysis}
    \resizebox{\columnwidth}{!}{%
    \begin{tabular}{ll ccccc}
    \toprule
        \textbf{Model} & \textbf{Metric} & \textbf{GQA} & \textbf{POPE} & \textbf{TextVQA} & \textbf{SQA} & \textbf{Avg.}  \\ 
        \midrule
        \multirow{3}{*}{InternVL (1B)} & QAA(Correct) & \textbf{89.0} & \textbf{82.9} & \textbf{89.8} & \textbf{65.4} & \textbf{81.8}  \\ 
        & QAA(Incorrect) & 79.2 & 80.7 & 71.7 & 45.9 & 69.4  \\ 
        \cline{2-7}
        & ACC & 10.0 & 59.9 & 30.8 & 18.9 & 29.9  \\
        \midrule
        \multirow{3}{*}{DeepSeek (1B)} & QAA(Correct) & \textbf{57.0} & \textbf{4.5} & \textbf{2.2} & 0.9 & \textbf{16.1}  \\ 
        & QAA(Incorrect) & 3.9 & 4.0 & 0.4 & \textbf{1.8} & 2.5  \\ 
        \cline{2-7}
        & ACC & 5.2 & 53.1 & 4.9 & 4.5 & 16.9  \\
        \midrule
        \multirow{3}{*}{Qwen (3B)} & QAA(Correct) & \textbf{66.2} & 63.8 & \textbf{63.9} & \textbf{55.0} & \textbf{62.2}  \\ 
        & QAA(Incorrect) & 64.6 & \textbf{64.0} & 63.7 & 44.7 & 59.3  \\ 
        \cline{2-7}
        & ACC & 45.7 & 80.0 & 66.6 & 61.2 & 63.4  \\
        \midrule
        \multirow{3}{*}{TinyLLaVA (3.1B)} & QAA(Correct) & \textbf{4.2} & \textbf{3.9} & \textbf{6.0} & 0 & \textbf{3.5}  \\ 
        & QAA(Incorrect) & 2.8 & 2.6 & 4.0 & \textbf{1.1} & 2.6  \\ 
        \cline{2-7}
        & ACC & 16.2 & 59.9 & 12.3 & 6.0 & 23.6   \\
        \midrule
        \multirow{4}{*}{BLIP (4B)} & QAA(Correct) & \textbf{81.0} & \textbf{89.6} & \textbf{64.5} & \textbf{68.9} & \textbf{76.0}  \\ 
        & QAA(Incorrect) & 65.4 & 86.9 & 45.8 & 44.1 & 60.5  \\ 
        \cline{2-7}
        & ACC & 40.3 & 79.6 & 50.7 & 28.6 & 49.8  \\
        \midrule
        \multirow{3}{*}{LLaVA (7B)} & QAA(Correct) & \textbf{37.4} & \textbf{34.6} & \textbf{35.0} & \textbf{34.5} & \textbf{35.4}  \\ 
        & QAA(Incorrect) & 33.2 & 32.6 & 31.5 & 22.6 & 30.0  \\ 
        \cline{2-7}
        & ACC & 18.4 & 62.1 & 21.5 & 3.6 & 26.4  \\
        \midrule
        \multirow{3}{*}{Doubao} & QAA(Correct) & \textbf{94.5} & \textbf{97.7} & \textbf{98.9} & \textbf{63.8} & \textbf{88.7}  \\ 
        & QAA(Incorrect) & 93.4 & 97.1 & 98.4 & 50.1 & 84.8  \\ 
        \cline{2-7}
        & ACC & 53.3 & 86.5 & 79.5 & 81.2 & 75.1 \\
        \bottomrule
    \end{tabular}
    }
\end{table}

\subsection{Few-shot Evaluation}
\label{app:few-shot-settings}

\begin{table*}[ht!]
\centering
\caption{Comparison of model performance on traditional VQA, VoQA \textit{pure visual-only} zero-shot, and few-shot settings across the VoQA benchmark. All results are reported in accuracy (\%). For few-shot results, each model shows the setting with the highest average accuracy.}

\label{tab:voqa_results}
\footnotesize
\begin{tabular}{lccccccc}
\toprule
\textbf{Model} & \textbf{Evaluation Setting} & \textbf{VQAv2} & \textbf{GQA} & \textbf{POPE} & \textbf{TextVQA} & \textbf{SQA} & \textbf{Avg.} \\
\midrule
\multirow{3}{*}{InternVL3-1B} 
& Traditional VQA   & \textbf{71.8}  & \textbf{52.1}  & \textbf{88.8}  & \textbf{71.8}  & \textbf{95.4}  & \textbf{76.0} \\
& VoQA zero-shot    & 5.6  & 3.2  & 53.9 & 12.8 & 19.5 & 19.0 \\
& VoQA few-shot     & \underline{20.2} & \underline{12.8} & \underline{59.5} & \underline{26.6} & \underline{26.0} & \underline{29.0} \\
\midrule
\multirow{3}{*}{Qwen2.5-VL-3B-Instruct} 
& Traditional VQA   & \textbf{82.0}  & \textbf{60.1}  & \textbf{88.2}  & \textbf{79.0}  & \textbf{84.9}  & \textbf{78.8} \\
& VoQA zero-shot    & 57.4 & 35.0 & 76.3 & 60.7 & 61.2 & 58.1 \\
& VoQA few-shot        & \underline{64.1} & \underline{41.4} & \underline{76.6} & \underline{64.4} & \underline{55.7} & \underline{60.4} \\
\midrule
\multirow{3}{*}{BLIP-3 (4B)} 
& Traditional VQA   & \textbf{81.7}   & \textbf{61.6}   & \textbf{87.0}  & \textbf{71.0}  & \textbf{89.7}  & \textbf{78.2} \\
& VoQA zero-shot    & 7.3  & 2.7  & 49.7 & 16.2 & 29.1 & 21.0 \\
& VoQA few-shot     & \underline{54.0} & \underline{35.8} & \underline{75.3} & \underline{51.3} & \underline{28.2} & \underline{48.9} \\
\bottomrule
\end{tabular}
\end{table*}

To help models learn the task pattern from examples, we incorporate few-shot settings into our prompt-engineering experiments.
We evaluate three open-source LVLMs (InternVL3-1B, Qwen2.5-VL-3B-Instruct, and BLIP-3 (4B)) in a few-shot setting, providing $k \in \{1, 2, 4, 8\}$ randomly sampled examples from the VoQA training dataset. Each example follows a \textit{JSON format} containing the question and answer. 

\begin{table}[!t]
    \centering
    \caption{\textit{InternVL3-1B} few-shot Answer Accuracy (ACC) and Question Alignment Accuracy (QAA) results. \textit{Correct} and \textit{Incorrect} indicate correctly and incorrectly answered samples, respectively. All results are reported in accuracy (\%).}
    \label{tab: internvl few-shot analysis}
    \resizebox{\columnwidth}{!}{%
    \begin{tabular}{ll cccccc}
    \toprule
        \textbf{Setting} & \textbf{Metric} & \textbf{VQAv2} & \textbf{GQA} & \textbf{POPE} & \textbf{TextVQA} & \textbf{SQA} & \textbf{Avg.}  \\ 
        \midrule
        \multirow{3}{*}{1-shot} & QAA(Correct) & / & \textbf{56.2} & \textbf{45.1} & \textbf{50.5} & \textbf{25.0} & \textbf{44.2}  \\ 
        & QAA(Incorrect) & / & 43.5 & 40.1 & 31.9 & 18.2 & 33.4  \\ 
        \cline{2-8}
        & ACC & 18.6 & 11.0 & 58.3 & 23.7 & 23.5 & 27.0  \\ 
        \midrule
        \multirow{3}{*}{2-shot} & QAA(Correct) & / & \textbf{53.5} & \textbf{42.4} & \textbf{51.2} & \textbf{31.0} & \textbf{44.5}  \\ 
        & QAA(Incorrect) & / & 45.3 & 39.7 & 34.9 & 18.1 & 34.5  \\ 
        \cline{2-8}
        & ACC & 20.2 & 12.8 & 59.5 & 26.6 & 26.0 & 29.0  \\ 
        \midrule
        \multirow{3}{*}{4-shot} & QAA(Correct) & / & \textbf{45.3} & \textbf{33.9} & \textbf{40.8} & \textbf{27.9} & \textbf{37.0}  \\ 
        & QAA(Incorrect) & / & 34.3 & 30.4 & 27.9 & 17.5 & 27.5  \\ 
        \cline{2-8}
        & ACC & 20.1 & 11.9 & 57.7 & 24.0 & 19.6 & 26.7 \\ 
        \midrule
        \multirow{3}{*}{8-shot} & QAA(Correct) & / & \textbf{38.3} & \textbf{29.8} & \textbf{31.9} & \textbf{15.7} & \textbf{28.9}  \\ 
        & QAA(Incorrect) & / & 28.9 & 27.0 & 22.0 & 13.8 & 22.9  \\ 
        \cline{2-8}
        & ACC & 19.2 & 11.8 & 57.1 & 21.0 & 14.5 & 24.7 \\ 
        \bottomrule
    \end{tabular}
    }
\end{table}

\paragraph{Few-shot settings.} 
We randomly sampled 10,000 instances from the VoQA training set to construct a demonstration pool. To ensure balanced coverage, the sampling process was stratified according to the proportion of samples from different sub-tasks in the training data. For each of the 134k evaluation samples, we randomly selected 1, 2, 4, or 8 examples from this pool as few-shot demonstrations. Once selected, the examples were fixed to ensure consistent comparison across models and reproducibility of results. 

Each selected example’s image, embedded question text, and corresponding answer were inserted into the following prompt template (shown below using the 2-shot setting as an example). In this template, the placeholders \texttt{<Ground Truth Question in Example x>} and \texttt{<Ground Truth Answer in Example x>} are replaced with the actual question and answer from the sampled examples.

\begin{table}[!t]
    \centering
    \caption{\textit{Qwen2.5-VL-3B-Instruct} few-shot Answer Accuracy (ACC) and Question Alignment Accuracy (QAA) results.}
    \label{tab: qwen few-shot analysis}
    \resizebox{\columnwidth}{!}{%
    \begin{tabular}{ll cccccc}
    \toprule
        \textbf{Setting} & \textbf{Metric} & \textbf{VQAv2} &\textbf{GQA} & \textbf{POPE} & \textbf{TextVQA} & \textbf{SQA} & \textbf{Avg.}  \\ 
        \midrule
        \multirow{3}{*}{1-shot} & QAA(Correct) & / & \textbf{67.3} & \textbf{63.7} & \textbf{71.3} & \textbf{42.3} & \textbf{61.1}  \\ 
        & QAA(Incorrect) & / & 63.9 & 61.8 & 69.6 & 30.5 & 56.4  \\ 
        \cline{2-8}
        & ACC & 64.1 & 41.4 & 76.6 & 64.4 & 55.7 & 60.4  \\ 
        \midrule
        \multirow{3}{*}{2-shot} & QAA(Correct) & / & \textbf{69.9} & \textbf{68.4} & \textbf{73.6} & \textbf{42.7} & \textbf{63.6}  \\ 
        & QAA(Incorrect) & / & 64.4 & 65.6 & 71.0 & 30.9 & 58.0  \\ 
        \cline{2-8}
        & ACC & 62.2 & 40.2 & 73.8 & 63.1 & 50.4 & 57.9  \\ 
        \midrule
        \multirow{3}{*}{4-shot} & QAA(Correct) & / & \textbf{66.3} & \textbf{67.7} & \textbf{69.7} & \textbf{44.0} & \textbf{61.9}  \\ 
        & QAA(Incorrect) & / & 58.6 & 63.6 & 66.2 & 29.1 & 54.4  \\ 
        \cline{2-8}
        & ACC & 59.5 & 38.5 & 71.1 & 61.7 & 46.7 & 55.5  \\ 
        \midrule
        \multirow{3}{*}{8-shot} & QAA(Correct) & / & \textbf{57.1} & \textbf{49.1} & \textbf{61.1} & \textbf{40.5} & \textbf{52.0}  \\ 
        & QAA(Incorrect) & / & 42.4 & 40.6 & 49.1 & 24.1 & 39.0  \\ 
        \cline{2-8}
        & ACC & 48.0 & 30.6 & 62.6 & 54.6 & 30.9 & 45.3 \\ 
        \bottomrule
    \end{tabular}
    }
\end{table}

\begin{table}[!t]
    \centering
    \caption{\textit{BLIP-3 (4B)} few-shot Answer Accuracy (ACC) and Question Alignment Accuracy (QAA) results.}
    \label{tab: blip few-shot analysis}
    \resizebox{\columnwidth}{!}{%
    \begin{tabular}{ll cccccc}
    \toprule
        \textbf{Setting} & \textbf{Metric} & \textbf{VQAv2} & \textbf{GQA} & \textbf{POPE} & \textbf{TextVQA} & \textbf{SQA} & \textbf{Avg.}  \\ 
        \midrule
        \multirow{3}{*}{1-shot} & QAA(Correct) & / & \textbf{66.8} & \textbf{63.1} & \textbf{57.6} & \textbf{25.6} & \textbf{53.3}  \\ 
        & QAA(Incorrect) & / & 53.2 & 58.5 & 48.5 & 13.7 & 43.5  \\ 
        \cline{2-8}
        & ACC & 53.6 & 35.4 & 77.3 & 51.1 & 26.6 & 48.8  \\ 
        \midrule
        \multirow{3}{*}{2-shot} & QAA(Correct) & / & \textbf{67.5} & \textbf{63.5} & \textbf{55.3} & \textbf{19.8} & \textbf{51.5}  \\ 
        & QAA(Incorrect) & / & 52.2 & 58.4 & 47.0 & 13.2 & 42.7  \\ 
        \cline{2-8}
        & ACC & 54.0 & 35.8 & 75.3 & 51.3 & 28.2 & 48.9 \\ 
        \midrule
        \multirow{3}{*}{4-shot} & QAA(Correct) & / & \textbf{56.5} & \textbf{40.7} & \textbf{38.8} & \textbf{4.5} & \textbf{35.1}  \\ 
        & QAA(Incorrect) & / & 27.3 & 26.2 & 22.9 & 4.7 & 20.3  \\ 
        \cline{2-8}
        & ACC & 37.7 & 26.2 & 64.3 & 42.4 & 26.3 & 39.4  \\ 
        \midrule
        \multirow{3}{*}{8-shot} & QAA(Correct) & / & \textbf{37.6} & \textbf{18.0} & \textbf{20.2} & \textbf{1.0} & \textbf{19.2}  \\ 
        & QAA(Incorrect) & / & 8.1 & 8.5 & 6.9 & 0.4 & 6.0  \\ 
        \cline{2-8}
        & ACC & 23.3 & 14.3 & 57.0 & 28.7 & 29.2 & 30.5 \\ 
        \bottomrule
    \end{tabular}
    }
\end{table}

\texttt{\\
Each image contains a question written inside it. Your task is to extract the question from the image and answer it accurately.\\
\\
Example 1:\\
Input: <image>\\
Output: \{"The question in the image": "<Ground Truth Question in Example 1>", "Answer": "<Ground Truth Answer in Example 1>"\}\\
\\
Example 2:\\
Input: <image>\\
Output: \{"The question in the image": "<Ground Truth Question in Example 2>", "Answer": "<Ground Truth Answer in Example 2>"\}\\
\\
Now answer the next one:\\
Input: <image>\\
Output:
}

\paragraph{Result analysis.} 
As shown in Table~\ref{tab:voqa_results}, the few-shot evaluation results on the VoQA benchmark show that few-shot examples help models better grasp the task pattern, yielding improvements over the zero-shot setting but still falling far short of performance on traditional VQA. The detailed results of each model on each sub-task can be found in Tables ~\ref{tab: internvl few-shot analysis}, ~\ref{tab: qwen few-shot analysis}, and ~\ref{tab: blip few-shot analysis}.

Consistent with our workflow analysis, all models exhibit higher Question Alignment Accuracy (QAA) for correctly answered samples across all few-shot configurations, underscoring that accurate question recognition remains the main factor for reliable reasoning. However, models still struggle to consistently identify and interpret embedded questions, leading to only modest overall gains.


\begin{table*}[h!]
    \centering
    \caption{ACC comparison of three fine-tuned models on the VoQA benchmark. All models share the same pre-trained backbone, and results are reported as accuracy (\%).}

    \label{tab: voqa-all-results}
    \footnotesize
    \begin{tabular}{lccccccc}
    \toprule
        \textbf{Base Model} & \textbf{Settings} & \textbf{VQAv2} & \textbf{GQA} & \textbf{POPE} & \textbf{TextVQA} & \textbf{SQA} & \textbf{Avg.}  \\ 
        \midrule
        \multirow{2}{*}{TinyLLaVA (1B)} 
        & VoQA zero-shot & 0.2 & 0.1 & 50.4 & 0 & 0 & 10.2  \\ 
         & Baseline-SFT & \textbf{63.4} & \textbf{45.3} & \textbf{75.5} & \textbf{32.5} & \textbf{24.4} & \textbf{48.2}  \\ 
         \midrule
        \multirow{2}{*}{InternVL (1B)} 
        & VoQA zero-shot & 5.5 & 3.2 & 53.9 & 12.8 & 19.5 & 19.0  \\ 
        & Baseline-SFT & \textbf{67.4} & \textbf{50.1} & \textbf{82.1} & \textbf{35.2} & \textbf{38.6} & \textbf{54.7}  \\ 
         \midrule
        \multirow{2}{*}{Qwen (2B)} 
         & VoQA zero-shot & 11.6 & 6.9 & 55.5 & 25.3 & 21.3 & 24.1  \\ 
        & Baseline-SFT & \textbf{75.9} & \textbf{56.9} & \textbf{85.9} & \textbf{68.5} & \textbf{54.3} & \textbf{68.3}  \\ 
         \bottomrule
    \end{tabular}
\end{table*}

\begin{table*}[!ht]
    \centering
    \caption{Evaluation of models fine-tuned on either VQA or VoQA using different strategies, all tested on traditional VQA benchmarks. Results are reported in accuracy (\%).}
    \label{tab: vqa-all-results}
    \footnotesize
    \begin{tabular}{l c c c c c c c}
    \toprule
    \textbf{Base Model} & \textbf{Training Setting} & \textbf{VQAv2} & \textbf{GQA} & \textbf{POPE} & \textbf{TextVQA} & \textbf{SQA} & \textbf{Avg.} \\
    \midrule
    
    \multirow{3}{*}{TinyLLaVA (1B)}   & VQA SFT & \underline{72.3}  & \textbf{55.8}  & \underline{86.6}  & \underline{42.1}  & \textbf{62.3}  & \textbf{63.8} \\
    & VoQA QA-SFT & 65.6  & 48.9  & 85.8  & 34.3 & 16.3  & 50.2 \\
    & VoQA QRA-SFT & \textbf{72.6}  & \underline{54.6}  & \textbf{87.1}  & \textbf{42.2}  & \underline{60.3}  & \underline{63.4} \\
    \midrule
    
    \multirow{3}{*}{InternVL (1B)} & VQA SFT  & \textbf{76.8}  & \textbf{56.6}  & \textbf{88.9}  & \textbf{69.0}  & \textbf{88.0}  & \textbf{75.8} \\
      & VoQA QA-SFT  & 74.6  & 50.9  & 85.7  & 62.5  & 85.3  & 71.8 \\
      & VoQA QRA-SFT  & \underline{76.3}  & \underline{56.1}  & \underline{88.0}  & \underline{64.5}  & \underline{85.4}  & \underline{74.1} \\
    \midrule
    \multirow{3}{*}{Qwen (2B)} & VQA SFT & \textbf{80.4} & \textbf{60.7} & \underline{87.8} & \textbf{76.0} & \textbf{82.6} & \textbf{77.5} \\
      & VoQA QA-SFT  &  62.1 & 50.7      & \textbf{88.3}      &    46.1   &  38.9     &     57.2  \\
      & VoQA QRA-SFT  &  \underline{70.6}  & \underline{58.6}      & 85.9      &      \underline{66.7} &  \underline{77.7}     &  \underline{71.9}     \\
    \bottomrule
    \end{tabular}%
\end{table*}

\newpage

\section{Question-Alignment Fine-Tuning}
\label{app: qa-like sft}

\begin{figure*}[!h]
    \centering
    \includegraphics[width=\linewidth]{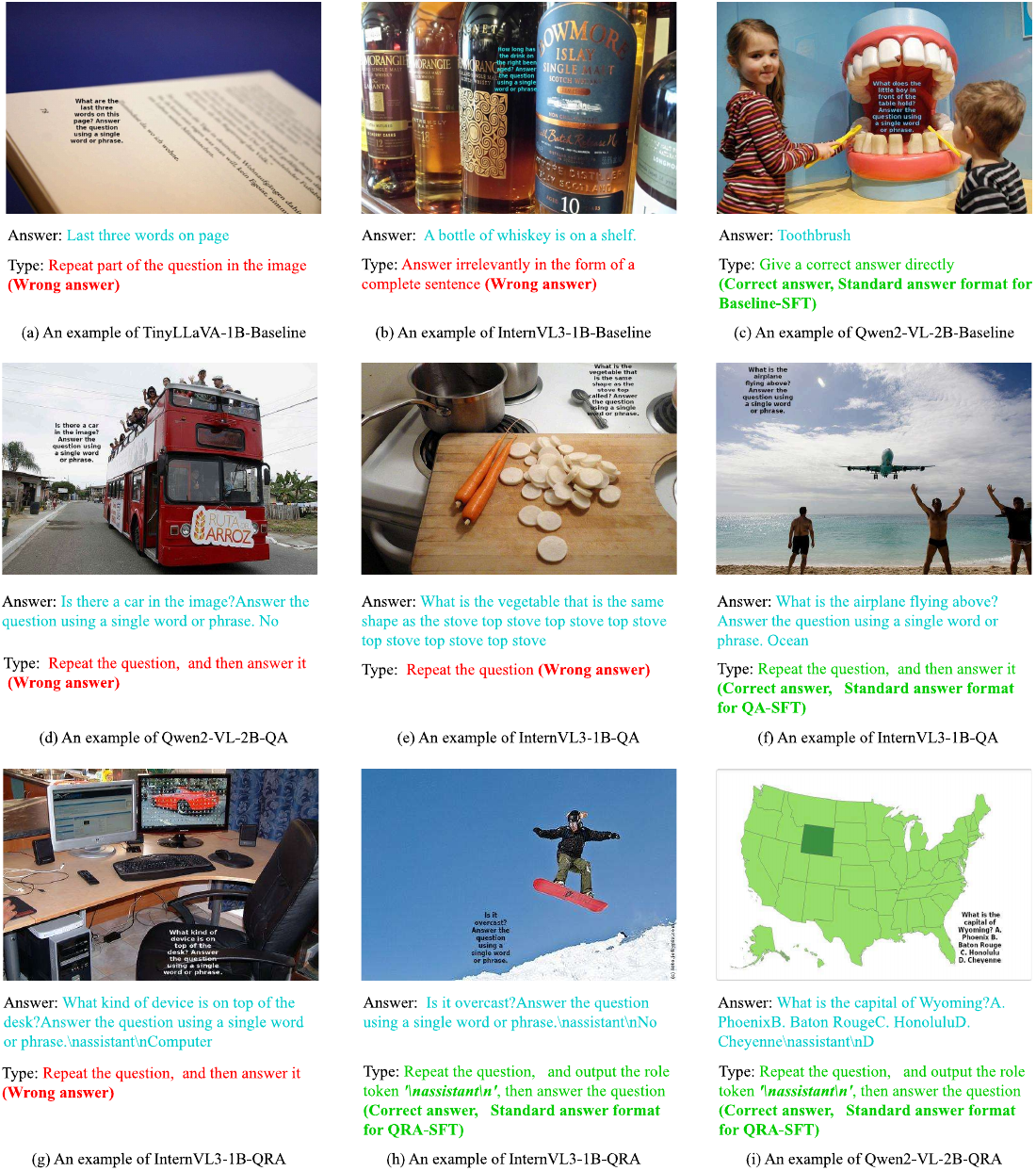}
    \caption{Nine representative case study examples illustrating the inference behavior of Baseline-SFT, QA-SFT, and QRA-SFT on the VoQA benchmark. The fine-tuning method used by each model is indicated as a suffix (-Baseline, -QA, -QRA) next to the model name. In each subfigure, the model’s predicted answer is shown in blue, while red and green indicate incorrect and correct answers, respectively, based on the \textit{Type} field. The correctness of each sample is explicitly marked.}

    \label{fig:example_case}
\end{figure*}

\subsection{Fine-Tuning Settings}
\label{app: sft setting}


All models are fine-tuned for one epoch with consistent hyperparameters, and all experiments are conducted on \(8 \times\) NVIDIA A100 (40GB) GPUs.
TinyLLaVA (1B) is trained on both the connector and language model using AdamW\cite{loshchilovdecoupled}, a cosine scheduler\cite{loshchilov2017sgdrstochasticgradientdescent} with 0.03 warmup, a learning rate of \(2 \times 10^{-5}\), and a batch size of 128.
For InternVL (1B) and Qwen (2B), only the language model is updated using LoRA\cite{hu2022lora} (rank 8), AdamW, and a cosine scheduler with 0.1 warmup; the learning rates are \(1 \times 10^{-5}\) and \(1.4 \times 10^{-5}\), respectively, with a batch size of 64.

\subsection{Case Study of Fine-Tuning Strategies}
\label{app: voqa examples}

Figure \ref{fig:example_case} presents representative examples illustrating the inference behaviors of Fine-tuning Strategies on the VoQA benchmark. Each subfigure includes the composite image input, the model’s prediction, a brief behavior analysis, and the corresponding model name. The examples cover diverse outcomes such as irrelevant responses, question repetition, and correct answers.

\subsection{Fine-tuning Results on VoQA and traditional VQA Benchmarks}
\label{app: complete sft results}

\begin{table}[!t]
    \centering
    \caption{Question Alignment Accuracy (QAA, \%) of models after QRA-SFT fine-tuning.  \textit{Correct} and \textit{Incorrect} indicate averages computed over correctly and incorrectly answered samples, respectively. Higher QAA indicates better recognition of visually embedded questions.}
    \label{tab:qra-sft-qaa}
    \resizebox{\columnwidth}{!}{%
    \begin{tabular}{lllcccccc}
    \toprule



        \textbf{Model} & \textbf{Results} & \textbf{GQA} & \textbf{POPE} & \textbf{TextVQA} & \textbf{SQA} & \textbf{Avg.} \\ 

        \midrule
        \multirow{2}{*}{TinyLLaVA} 
        & Correct   & \textbf{95.8} & \textbf{98.6} & \textbf{97.2} & \textbf{87.2} & \textbf{94.7} \\ 
        & Incorrect & 93.1 & 95.1 & 93.9 & 73.6 & 88.9 \\ 

        \midrule
        \multirow{2}{*}{InternVL} 
        & Correct   & \textbf{96.6} & \textbf{98.3} & \textbf{97.6} & \textbf{89.7} & \textbf{95.5} \\ 
        & Incorrect & 96.1 & 97.6 & 93.6 & 57.9 & 86.3 \\ 

        \midrule
        \multirow{2}{*}{Qwen} 
        & Correct   & \textbf{97.7} & \textbf{98.7} & \textbf{98.6} & \textbf{92.8} & \textbf{96.9} \\ 
        & Incorrect & 97.5 & 98.8 & 91.7 & 73.8 & 90.4 \\ 

        \bottomrule
    \end{tabular}
    }
\end{table}

\paragraph{Complete Results of VoQA ACC.}
Table~\ref{tab: voqa-all-results} shows that Baseline-SFT yields consistent performance gains over the zero-shot setting across all VoQA sub-tasks.

\paragraph{Complete Results of VQA ACC.}
From the results in Table~\ref{tab: vqa-all-results}, QRA-SFT demonstrates clear advantages over QA-SFT on most VQA sub-tasks and even surpasses VQA-SFT in several cases. These findings highlight the strength of QRA-SFT in maintaining the traditional VQA input format while effectively aligning visual questions during VoQA-oriented fine-tuning.

\paragraph{Complete Results of VoQA QAA.}
As summarized in Table~\ref{tab:qra-sft-qaa}, the average QAA of correct samples consistently exceeds that of incorrect ones across all sub-tasks, indicating that accurately detecting the embedded question is a crucial prerequisite for reliable answer generation.

\begin{table*}[!ht]
    \centering
    \caption{
    Impact of including or excluding the role token during inference on traditional VQA benchmarks for both VoQA SFT variants. \textit{w/ Role} and \textit{w/o Role} indicate whether the role token is explicitly provided in the fixed input template at training or inference time, i.e., whether the model receives this token immediately before generating the answer. Results are reported in accuracy (\%).
    }
    \footnotesize
    \label{tab: vqa}
    \begin{tabular}{llllcccccc}
    \toprule
        \textbf{Model} & \textbf{SFT} & \textbf{Train} & \textbf{Inference} & \textbf{VQAv2} & \textbf{GQA} & \textbf{POPE} & \textbf{TextVQA} & \textbf{SQA} & \textbf{Avg.} \\ 
        \midrule
        \multirow{4}{*}{TinyLLaVA (1B)}& \multirow{2}{*}{QA-SFT} & \multirow{2}{*}{w/ Role} & w/ Role & \textbf{65.6} & \textbf{48.9} & \textbf{85.8} & \textbf{34.3} & \textbf{16.3} & \textbf{50.2}  \\ 
        & & & w/o Role & 20.3 & 7.7 & 50.5 & 16.4 & 3.1 & 19.6 \\ 
        \cline{2-10}
        & \multirow{2}{*}{QRA-SFT} & \multirow{2}{*}{w/o Role} & w/ Role & \textbf{72.6} & \textbf{54.6} & \textbf{87.1} & \textbf{42.2} & 60.3 & \textbf{63.4}  \\ 
        & & & w/o Role & 69.6 & 52.2 & \textbf{87.1} & 40.7 & \textbf{60.4} & 62.0 \\ 

        \midrule

        \multirow{4}{*}{InternVL (1B)} & \multirow{2}{*}{QA-SFT} & \multirow{2}{*}{w/ Role} & w/ Role & \textbf{74.6} & \textbf{50.9} & \textbf{85.7} & \textbf{62.5} & \textbf{85.3} & \textbf{71.8}  \\ 
        & & & w/o Role & 12.2 & 4.4 & 50.6 & 3.4 & 48.8 & 23.9  \\ 
        \cline{2-10}
        & \multirow{2}{*}{QRA-SFT} & \multirow{2}{*}{w/o Role} & w/ Role & \textbf{76.3} & \textbf{56.1} & \textbf{88.0} & \textbf{64.5} & \textbf{85.4} & \textbf{74.1}  \\ 
        & & & w/o Role & 74.6 & 54.5 & 85.9 & 63.7 & 81.6 & 72.0 \\

        \midrule
        
        \multirow{4}{*}{Qwen (2B)} & \multirow{2}{*}{QA-SFT} & \multirow{2}{*}{w/ Role} & w/ Role &   \textbf{62.1} &  \textbf{50.7} & \textbf{88.3} & \textbf{46.1} & \textbf{38.9} &  \textbf{57.2}  \\
        & & & w/o Role & 0 & 0 & 51.1 & 0 & 0.6 & 10.3 \\
        \cline{2-10}
        & \multirow{2}{*}{QRA-SFT} & \multirow{2}{*}{w/o Role} & w/ Role  &  \textbf{70.6}  & \textbf{58.6}      & \textbf{85.9}      &   \textbf{66.7} &  \textbf{77.7}     &  \textbf{71.9}     \\
        & & & w/o Role & 63.4 & 47.8 & 79.1 & 37.4 & 49.8 & 55.5 \\
        
        \bottomrule
    \end{tabular}
\end{table*}

\subsection{Influence of Role Tokens in VQA Inference Templates}
\label{sec: role token in VQA}

\paragraph{Purpose.}
Since QA-SFT alters the input sequence format originally used in traditional VQA, we examine how the inclusion of the role token (i.e., the \textit{ASSISTANT:} token shown in Figure~\ref{fig: sft}) influences model behavior during VQA inference. Specifically, we analyze whether retaining this token helps preserve VQA performance and how its presence or absence affects QRA-SFT.

\paragraph{Findings.}
Table~\ref{tab: vqa} shows that including the role token generally improves the performance of both SFT variants, suggesting that the token serves as a useful cue that prompts the model to begin generating an answer in traditional VQA settings. Nevertheless, QRA-SFT without this token remains highly competitive, indicating that the model implicitly learns to anticipate this token during VoQA fine-tuning and thus can still maintain robust question–answer alignment even without explicitly providing the structural cue.

\subsection{Exploring Variants of Question-Alignment Fine-Tuning}
\label{sft variants}

\begin{figure}[htbp]
\centering
\includegraphics[width=1\linewidth]{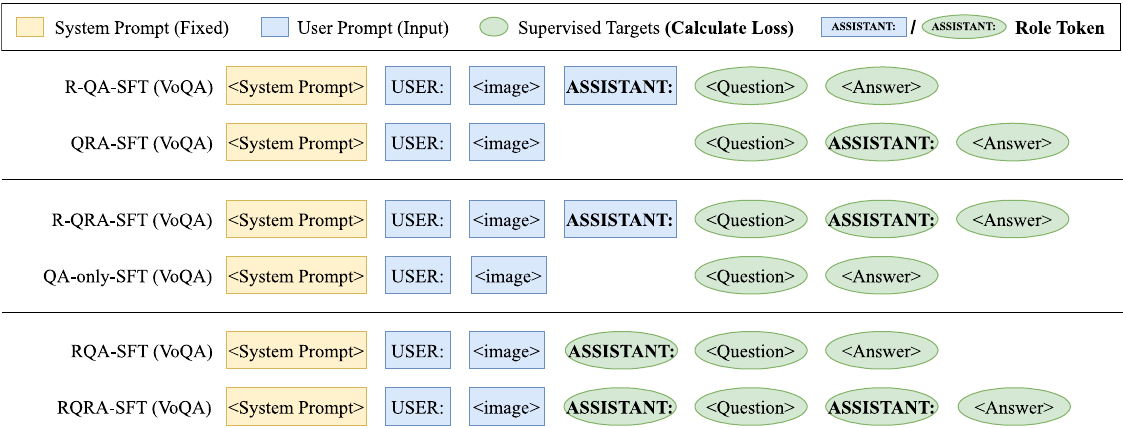}

\caption{
Comparison of six supervised fine-tuning strategies on the VoQA dataset. All methods are our proposed VoQA-specific approaches that first align the visually grounded question before generating the answer. The first two correspond to the main methods discussed in the paper: QA-SFT (denoted here as R-QA-SFT for consistency across variants) and QRA-SFT. The middle two are their variants, obtained by adding or removing the role token relative to the original methods. The last two correspond to variants of R-QA-SFT and R-QRA-SFT in which the model is required to explicitly predict the role token at the beginning of the output sequence.
}
\vspace{-10pt}
\label{fig: sft-appendix}
\end{figure}

\subsubsection{Analyzing the Effect of Role Tokens in Question-Alignment Fine-Tuning}

Since QA-SFT and QRA-SFT differ in both the placement of role tokens and whether these tokens appear in the supervised target, we further investigate how role-token positioning and supervision design influence model behavior.

\paragraph{R-QRA-SFT.}
As illustrated in Figure~\ref{fig: sft-appendix} (Line 3), \textit{Role-guided Question–Role–Answer Supervised Fine-Tuning} (R-QRA-SFT) unifies the strengths of QA-SFT and QRA-SFT under a consistent structure. The input includes a role token that identifies the model as an \textit{ASSISTANT}, serving as a cue to initiate answer generation. Meanwhile, the supervised target expands to the complete \textit{Question–Role–Answer} sequence, making the boundary between question interpretation and answer production explicit. This design enforces structured reasoning while preserving tight visual–textual alignment.

\paragraph{QA-only-SFT.}
To isolate the effect of explicit role cues, we introduce \textit{Question–Answer only Supervised Fine-Tuning} (QA-only-SFT, Figure~\ref{fig: sft-appendix}, Line 4), where both the input and output sequences omit the \textit{ASSISTANT} token, while the model still predicts the full question–answer pair. This variant tests whether removing role tokens weakens the model’s ability to align visually embedded questions with correct answers.

\begin{table}[!ht]
    \centering
    \caption{Comparison between five fine-tuning strategies on the VoQA benchmark. All results are reported in accuracy (\%).}
    \label{tab: qa-like sft2}
    \resizebox{\columnwidth}{!}{%
    \begin{tabular}{llcccccc}
    \toprule
        \textbf{Model} & \textbf{VoQA SFT} & \textbf{VQAv2} & \textbf{GQA} & \textbf{POPE} & \textbf{TextVQA} & \textbf{SQA} & \textbf{Avg.}  \\ 
        \midrule
        \multirow{5}{*}{TinyLLaVA (1B)} 
        & Baseline-SFT & 63.4 & 45.3 & 75.5 & 32.5 & 24.4 & 48.2  \\
        &  QA-SFT & \textbf{70.0} & \textbf{49.8} & \textbf{84.1} & 37.4 & 36.9 & \textbf{55.6}  \\ 
        &  QRA-SFT & 69.6 & 49.5 & 83.8 & 37.1 & \textbf{38.2} & \textbf{55.6} \\
        & R-QRA-SFT & 69.7 & 49.6 & 84.0 & 37.4 & 37.3 & \textbf{55.6} \\
        & QA-only-SFT & 69.2 & 48.7 & 83.8 & \textbf{37.5} & 36.1 & 55.1  \\  
        \midrule
        
        \multirow{5}{*}{InternVL (1B)} 
        & Baseline-SFT & 67.4 & 50.1 & 82.1 & 35.2 & 38.6 & 54.7  \\
        &  QA-SFT & \textbf{73.2} & 53.0 & 86.6 & 53.7 & 42.5 & 61.8  \\ 
        &  QRA-SFT & 72.6 & 52.7 & 85.8 & \textbf{56.3} & 49.1 & 63.3  \\ 
        & R-QRA-SFT & 72.5 & \textbf{53.2} & \textbf{87.3} & 56.2 & 50.3 & \textbf{63.9}  \\ 
        & QA-only-SFT & 69.5 & 51.2 & 86.3 & 53.5 & \textbf{50.5} & 62.2  \\
        
        \midrule
        \multirow{5}{*}{Qwen (2B)} 
        & Baseline-SFT & 75.9 & 56.9 & 85.9 & 68.5 & 54.3 & 68.3  \\ 
        &  QA-SFT & 79.2 & 60.1 & 87.6 & 70.8 & 60.6 & 71.7  \\ 
        &  QRA-SFT & 78.1 & 60.5 & 88.0 & 70.8 & 60.5 & 71.6  \\
        & R-QRA-SFT & 78.0 & 60.5 & \textbf{88.1} & 71.9 & \textbf{63.9} & \textbf{72.5} \\ 
        & QA-only-SFT & \textbf{79.3} & \textbf{60.6} & 87.9 & \textbf{72.3} & 58.1 & 71.6  \\
        \bottomrule
    \end{tabular}
    }
\end{table}

\begin{table}[!b]
    \centering
    \caption{Comparison of two fine-tuning strategies on the VoQA benchmark using the \textit{InternVL3-1B} model, evaluating their performance with and without providing the role token during inference. All results are reported in accuracy (\%).}
    \resizebox{\columnwidth}{!}{%
    \begin{tabular}{lll cccccc}
    \toprule
        \textbf{SFT} & \textbf{Train} & \textbf{Inference} & \textbf{VQAv2} & \textbf{GQA} & \textbf{POPE} & \textbf{TextVQA} & \textbf{SQA} & \textbf{Avg.} \\
        \midrule
        \multirow{2}{*}{RQA} & \multirow{2}{*}{w/o Role} & w/ Role & \textbf{71.9} & \textbf{52.0} & \textbf{85.5} & \textbf{51.6} & \textbf{44.1} & \textbf{61.0}  \\
        & & w/o Role & 37.0 & 22.5 & 68.1 & 29.8 & 25.6 & 36.6 \\
        \midrule
        \multirow{2}{*}{RQRA} & \multirow{2}{*}{w/o Role} & w/ Role & \textbf{72.8} & \textbf{53.1} & \textbf{86.9} & \textbf{56.8} & \textbf{51.3} & \textbf{64.2}  \\
        & & w/o Role & 5.2 & 4.2 & 52.0 & 6.6 & 30.8 & 19.8 \\
        \bottomrule
    \end{tabular}
    }
\end{table}

\paragraph{Result Analysis.}
Table~\ref{tab: qa-like sft2} shows that QA-only-SFT performs slightly worse than R-QRA-SFT, though their results remain highly comparable. This indicates that role tokens provide only a weak segmentation cue in this setting. When compared with QA-SFT and QRA-SFT introduced in Section~\ref{sec:qa-like sft}, all four question-aligned fine-tuning strategies exhibit similar performance and consistently surpass Baseline-SFT by a substantial margin. Overall, the results confirm that precise question alignment is the primary factor driving the model’s success on the VoQA task, while the presence of role tokens plays a relatively minor role.

\subsubsection{Effect of Predicting the Initial Role Token}

\paragraph{Two variants.}
To further understand how role tokens influence question-aligned fine-tuning, we analyze whether models can autonomously infer their role instead of relying on an explicit role token placed at the beginning of the output sequence, which is the default design in both R-QA-SFT and R-QRA-SFT. Building on these strategies, we extend the supervised targets to include the initial role token, yielding \textit{Role–Question–Answer Supervised Fine-Tuning} (RQA-SFT, Figure~\ref{fig: sft-appendix}, Line 5, from R-QA-SFT) and \textit{Role–Question–Role–Answer Supervised Fine-Tuning} (RQRA-SFT, Figure~\ref{fig: sft-appendix}, Line 6, from R-QRA-SFT).

\paragraph{Findings.}
Experiments on \textit{InternVL} models reveal a consistent pattern. When the role token is not provided during inference, both variants show degraded performance because the model struggles to predict the token reliably. However, once the role token is supplied as an input cue, performance improves substantially. The findings demonstrate that VoQA performance is determined primarily by whether the model initiates generation with accurate alignment to the visually embedded question, while the presence or prediction of a role token has no substantive impact. In other words, successful VoQA performance depends far more on accurate question alignment at the first step than on predicting an explicit \textit{ASSISTANT} role token.

\end{document}